# Predicting and Mapping of Soil Organic Carbon Using Machine Learning Algorithms in Northern Iran

**Mostafa Emadi [1], Ruhollah Taghizadeh-Mehrjardi [2,3], Ali Cherati [4], Majid Danesh [1], Amir Mosavi [5,6,7,*] and Thomas Scholten [2,8,9]**

1 Department of Soil Science, College of Crop Sciences, Sari Agricultural Sciences and Natural Resources University, Sari 4818168984, Iran; mostafa.emadi@sanru.ac.ir (M.E.); m.danesh@sanru.ac.ir (M.D.)
2 Department of Geosciences, Soil Science and Geomorphology, University of Tübingen, 72070 Tübingen, Germany; ruhollah.taghizadeh-mehrjardi@mnf.uni-tuebingen.de (R.T.-M.); thomas.scholten@uni-tuebingen.de (T.S.)
3 Faculty of Agriculture and Natural Resources, Ardakan University, Ardakan 8951656767, Iran
4 Soil and Water Research Department, Mazandaran Agricultural and Natural Resources Research and Education Center, AREEO, Sari 4849155356, Iran; a.cherati@areeo.ac.ir
5 Faculty of Civil Engineering, Technische Universität Dresden, 01069 Dresden, Germany
6 School of the Built Environment, Oxford Brookes University, Oxford, UK
7 Department of Informatics, J. Selye University, 94501 Komarno, Slovakia
8 CRC 1070 Ressource Cultures, University of Tübingen, 72070 Tübingen, Germany
9 DFG Cluster of Excellence "Machine Learning", University of Tübingen, 72070 Tübingen, Germany
* Correspondence: amir.mosavi@mailbox.tu-dresden.de

**Abstract:** Estimation of the soil organic carbon (SOC) content is of utmost importance in understanding the chemical, physical, and biological functions of the soil. This study proposes machine learning algorithms of support vector machines (SVM), artificial neural networks (ANN), regression tree, random forest (RF), extreme gradient boosting (XGBoost), and conventional deep neural network (DNN) for advancing prediction models of SOC. Models are trained with 1879 composite surface soil samples, and 105 auxiliary data as predictors. The genetic algorithm is used as a feature selection approach to identify effective variables. The results indicate that precipitation is the most important predictor driving 14.9% of SOC spatial variability followed by the normalized difference vegetation index (12.5%), day temperature index of moderate resolution imaging spectroradiometer (10.6%), multiresolution valley bottom flatness (8.7%) and land use (8.2%), respectively. Based on 10-fold cross-validation, the DNN model reported as a superior algorithm with the lowest prediction error and uncertainty. In terms of accuracy, DNN yielded a mean absolute error of 0.59%, a root mean squared error of 0.75%, a coefficient of determination of 0.65, and Lin's concordance correlation coefficient of 0.83. The SOC content was the highest in udic soil moisture regime class with mean values of 3.71%, followed by the aquic (2.45%) and xeric (2.10%) classes, respectively. Soils in dense forestlands had the highest SOC contents, whereas soils of younger geological age and alluvial fans had lower SOC. The proposed DNN (hidden layers = 7, and size = 50) is a promising algorithm for handling large numbers of auxiliary data at a province-scale, and due to its flexible structure and the ability to extract more information from the auxiliary data surrounding the sampled observations, it had high accuracy for the prediction of the SOC base-line map and minimal uncertainty.

## 1. Introduction

Soil organic carbon (SOC) is central to soil health as it plays a significant role in soil aggregation, water holding capacity, cation/anion exchangeability, and nutrient availability, which promotes plant growth. SOC can potentially affect both soil ecosystems and crop productivity due to its several critical roles in soil functioning. Globally, the amount of carbon in the upper one meter of soil is about three and two times higher than the amount of carbon found in the biosphere and atmosphere, respectively [1–3]. Therefore, the contribution of SOC to the global C cycle by sequestering terrestrial C is of great importance. Changes in SOC pools induced by soil management and land cover changes affect global warming and, in turn, can significantly influence soil physical, chemical, and biological properties [4–6]. As SOC is a good indicator of environmental quality [7], high-quality maps of the spatial distribution of SOC can provide base-line data for SOC turnover and sequestration for C management strategies at the province-scale. The spatial variability of SOC at the field to the regional scale is highly related to the soil forming-factors including the climate (precipitation and temperature); organisms (vegetation and human), relief (terrain attributes), parent materials, and time [8].

Due to the global importance of SOC, digital soil mapping (DSM) approaches have become more focused on SOC mapping in the last decade [4,9–12]. DSM describes the spatial variation of SOC by taking the relations between SOC and environmental auxiliary variables into account [13–15]. The auxiliary variables correlated with SOC are often obtained from digital elevation models [11,16], remotely sensed data [16,17] and climatic data [18,19]. By using remotely sensed imagery and easy accessibility of climatic and digital elevation model (DEM) data, the application of machine learning (ML) techniques for predicting SOC is significantly increased [11,17,20].

Numerous ML algorithms have been applied in DSM for SOC prediction including artificial neural networks (ANNs) [21–24], genetic programming [25], support vector regression (SVR) [23,25,26], multivariate adaptive regression splines [27,28], Cubist [9,29,30], boosted regression tree [16,31,32], and random forest (RF) [9,16,19,23,32–34]. In most cases, these approaches were much more accurate than linear and geostatistical methods due to the higher ability to get a lot more information for unsampled points by investigating nonlinear relationships between SOC and environmental auxiliary variables.

Soils, especially SOC contents, which are the result of the actions and interactions of many different processes and factors, vary from place to place with high complexity [35]. Thus, to predict the behaviors and properties of such a complex environment, classical ML may encounter problems [10,12,34,36]. A new approach that has received considerable attention as a sophisticated learning algorithm with substantial learning capability and high performance is deep learning (DL) [37]. Recently, deep neural networks (DNNs) based on DL approaches have been proposed for overcoming the shortages arising from the traditional ANNs [37,38], by adding more complexity (deep) into the conventional models. This provides better learning capabilities to reveal the complexity underlying the data, and thus results in a higher accuracy of the trained model [39]. The hierarchical structure and high learning capacity make DNN models quite flexible and adaptable for a wide variety of highly complex problems such as SOC prediction [11,39,40]. The DNNs have recently been used for the prediction of soil properties [40–42] and particularly for SOC prediction [43,44]. Xu et al. [43], for instance, indicated that the DNN method had a high performance for the prediction of SOC with the effective abstraction of complex covariates for learning by using visible and near-infrared soil spectra. DNNs are more complex and need further parametrization but severely depend on the size of the training dataset.

To eliminate the multicollinearity of variables and exclude unimportant and redundant auxiliary variables, numerous feature selection techniques have been developed for DSM including. Particle swarm optimization [45], the genetic algorithm (GA) [25,32,46], hybrid GA-artificial neural network [47], parallel GA [48], and the artificial bee colony algorithm [36] are among the notable feature selection techniques. Such variable selection techniques can simplify modeling by lowering the number of input variables and potentially improving the accuracy of soil predictions. There is no universal feature selection method to reduce the number of covariates in the pool presented to an ML algorithm. For instance, Behrens et al. [49] compared the two most common approaches for the selection of covariates, namely supervised and unsupervised, and found that the supervised feature selection approach was superior because the soil classes were predicted more accurately. Taghizadeh-mehrjardi et al. [50] explored the effect of the reduction in dimension of feature space with ant colony optimization (ACO) and correlation-based feature selection (CFS) on the accuracy of prediction of spatial models for each particle size fraction. In this study, we decided to implement the GA, one of the most advanced algorithms for feature selection [46]. GA can manage the datasets with many features and do not need specific knowledge about the problem parallelizing easily in computer clusters [46,51,52].

Although many ML algorithms have been developed for the prediction of soil properties, the development of site-specific techniques is necessary for enhancing the quality of thematic soil maps [53]. However, there is no best worldwide predictive algorithm for SOC mapping given that the accuracy level of SOC predictions is highly related to the local geographic attributes of the study area [54], the sampling size [9,55] and the selected auxiliary variates [14,19,56].

Mazandaran province, northern Iran, is located on the southern coast of the Caspian Sea. There is a descending precipitation gradient from the west to east across the region, leading to a diversity of soil moisture regime (SMR) and soil temperature regime (STR) classes [57]. Due to the changes in SOC contents in northern Iran caused by the human activities and natural attributes (landslide, flooding, depression) [58,59], the existence of a high-quality SOC prediction map with known uncertainty in the Mazandaran province is crucial. This provides a base-line map for further temporal monitoring of SOC at the province-scale. Despite the known advantages of feature selection, there have been no insights into the important variables for SOC prediction in northern Iran given the different predominant climatic and soil-forming conditions. Therefore, due to the lack of an SOC base-line distribution map in Mazandaran province, the objectives of this research were (1) to determine the important auxiliary variables driving the SOC contents in the province using GA as a popular automatic method for feature selection, (2) to test the performance of six ML algorithms fed with GA-selected auxiliary variables and (3) to predict the spatial distribution of SOC for mapping with associated uncertainty and (4) to compare SOC contents in different geological units, soil classes and land uses in Mazandaran province.

## 2. Materials and Methods

### *2.1. Study Area*

This research was conducted in Mazandaran province, northern Iran. The region is located at a longitude of 50°31′21″ E to 53°56′52″ E and latitude of 36°38′06″ N to 36°54′59″ N and covers an area of 2,388,179 ha. It borders the Caspian Sea in the north and the Alborz Mountain range in the south (Figure 1). Most of the province is covered by dense, moderate, and low-density forest with each forest type covering 39%, 4%, and 2% of the total area, respectively. There are several kinds of cultivated lands in the study area. Paddy fields are the most common agricultural land use with about 210,000 ha, and orchards cover about 90,000 ha. Based on the De Martonne climate classification, the western, central, eastern, and mountainous parts of the province have very humid, humid, Mediterranean, and semihumid climates, respectively. The mean annual temperature ranges from 18 °C on the coastal plain to below 8 °C in the highlands. There is a gradient of the decreasing precipitation from the west (around 1400 mm) to the east direction (around 450 mm) leading to the diversity of soil moisture regime (SMR) and soil temperature regime (STR) classes across the province. The xeric SMR class covers the largest area in the province, followed by the udic and aquic classes, while thermic (66%) is the most abundant STR followed by mesic (33%) and cryic (1%) [57]. The variation of elevation ranges from the Caspian Sea coastal areas with elevations <−5 m to more than 3000 m above sea level in the highlands of the Alborz Mountain range. Five USDS soil taxonomy orders including Mollisols, Entisols Inceptisols, Alfisols, and Ultisols were in Mazandaran with a total of 12 suborders. Mollisols are the most dominant soils forming on different landforms with mollic epipedons. Mollisols have four main suborders including the aquolls, rendolls, udolls, and xerolls mostly distinguishing based on soil moisture regime classes except for rendolls that have an epipedon with less than 50 cm thickness overlies on a highly calcareous horizon. Alfisols are characterized by a clay enriched endopedon with two main suborders, i.e., aqualfs and udalfs. Ultisols with a single suborder, i.e., Humults occurring in a very limited area in Mazandaran province with leached soils under native forest vegetation. Entisols with three main suborders, i.e., Orthents, Fluvents, and Aquents are mostly used for paddy cultivation in the study area. Inceptisols in Mazandaran province have a weakly developed B horizon with two main suborders, i.e., Aquept and Xerepts.

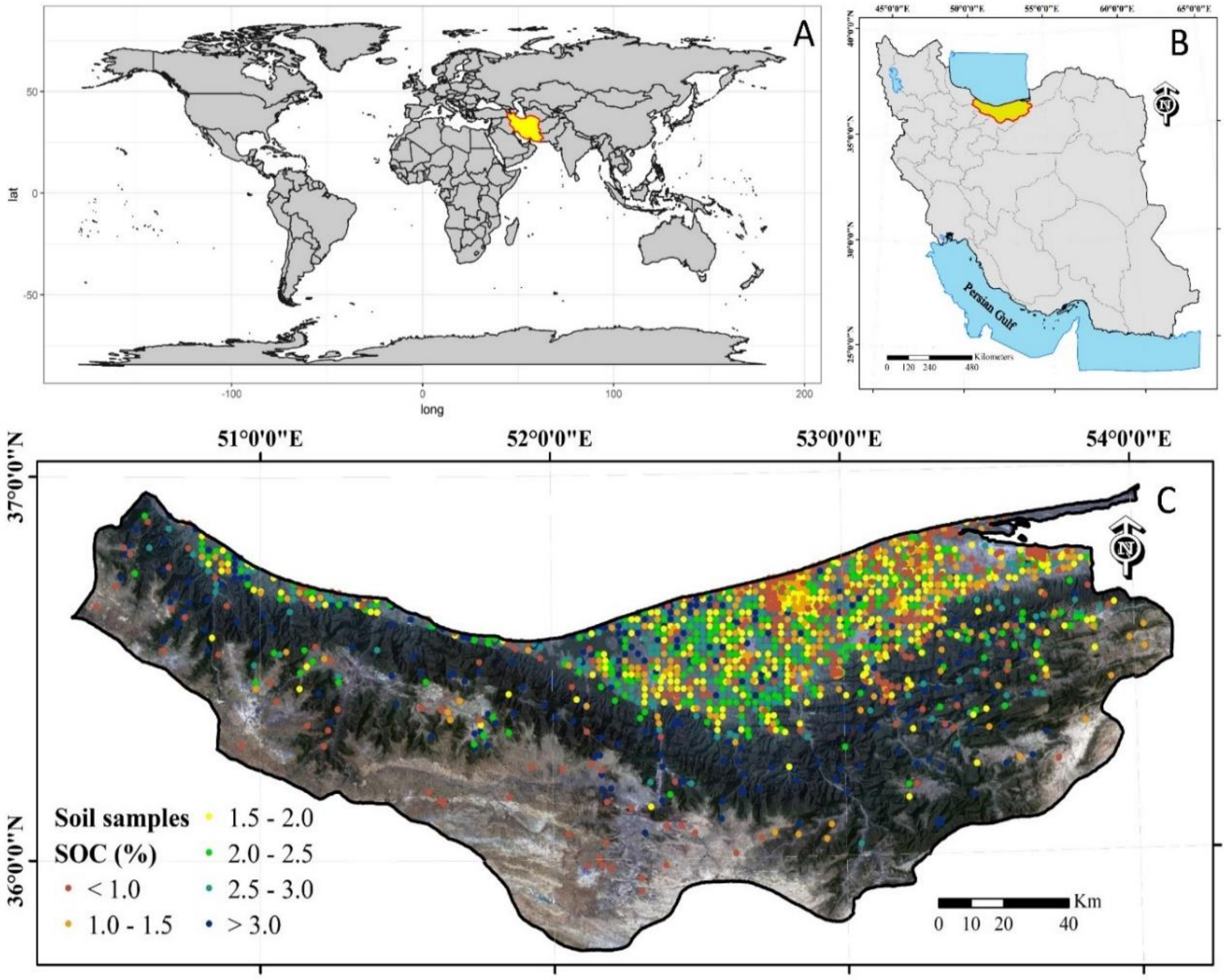

**Figure 1.** (**A**) Geographical position of Iran on the world map, (**B**) geographical position of the study area in Iran, and (**C**) spatial distribution of soil samples. The background is a false-color composite image derived from Landsat.

### 2.2. Soil Data

The total dataset for SOC mapping is 1879 composite surface soil samples from two main sources (Figure 1). More than half of the data (1055 samples) were derived from five Master of Science (M.Sc.) research projects in the soil science department at Sari Agricultural Sciences and Natural Resources University (SANRU) [60–64]. These samples were collected using a simple random sampling scheme mostly in uncultivated areas. The rest of the dataset comes from soil surveys performed by the Agricultural Research Education and Extension Organization (AREEO) and the Ministry of Jahad-e-Keshvarzi in Sari, Northern Iran. These samples were mostly collected in cultivated areas spread across the province using a grid sampling scheme with a 2000 m grid interval. Each composite soil sample is collected within a 20 m radius surrounding each of the sampling points with at least 10 subsamples (cores). Samples were collected from a depth of 0–20 cm, and their geographical coordinates were recorded with a global positioning system (GPS) device. After air-drying and passing through a 2 mm sieve, the content of SOC was measured by the wet oxidation procedure outlined by the Walkley and Black [65]. Figure 1 shows the spatial distribution of the sampling sites across Mazandaran province.

*2.3. Auxiliary Variables*

The full set of 105 predictor variables initially considered is given in Table A1 in the Appendix A. The auxiliary data included variables derived from remotely sensed imagery (60 variables from Landsat 8 and MODerate-resolution Imaging Spectroradiometer, MODIS), terrain attributes (30 variables), climatic data (10 variables), and five categorical data (e.g., soil map and land use map).

The high contribution of SOC to soil color can be detected by the spectral signature of remotely sensed data. The 60 environmental auxiliary variables derived from satellite imagery were developed based on the median values of 8 satellite images of the Landsat 8 Operational Land Imager taken from 2012 to 2016 to coincide with the dates of soil sampling. Following radiometric, geometric, and atmospheric corrections digital numbers for the blue (B1), green (B2), red (B3), near-infrared (B4), and shortwave IR-2 bands (B6) were extracted. Several indices were then calculated: the normalized difference vegetation index (NDVI), enhanced vegetation index (EVI), combined spectral response index (COSRI), land surface water index (LSWI), brightness index (BI), and other indices with a spatial resolution of 30 m. The variables derived from MODIS imagery had a spatial resolution of 250 m. These included median values of two spectral reflectance bands: red (645 nm) and near-infrared (858 nm) and the EVI, NDVI, and other indices. The daytime and nighttime land surface temperature with a 1 km resolution were also derived from MODIS data. Overall, 60 auxiliary variables were derived from Landsat 8 and MODIS data.

Thirty terrain attributes were derived from a DEM [66]. The DEM was obtained from shuttle radar topography mission terrain (SRTM) data with 30 m grid cells. Terrain attributes, namely slope, aspect, elevation, length and steepness (LS) factor, valley depth, openness, catchments area, catchment slope, plane curvature, topographic wetness index (TWI), channel networks base level (CNBL), distance to channel networks, the multiresolution valley bottom flatness index (MrVBF) and other indices [67], were calculated using SAGA GIS.

Climatic factors have high potential to explain large parts of the variation of soil properties in the northern part of Iran, due to the high degree of spatial variability in Mazandaran province. In this study, 10 climatic variables were obtained using WorldClim. WorldClim version 2 [68] contains reliable temperature and precipitation data at a spatial resolution of 1000 m. Categorical predictor variables were derived from five choropleth maps, which were compiled at different cartographic scales, e.g., soil map and land use map [69].

Figure 2 shows the spatial distribution of some auxiliary variables related to SOC including precipitation, NDVI, MrVBF, and land use. The precipitation decreases from the north-west to the north-east in the province, especially in shoreline areas (Figure 2). The southern parts of the province have lower precipitation compared with the northern region where the Caspian Sea shoreline is located. The NDVI values range from −0.5 to more than 0.8, indicating a high diversity of vegetation cover spread across the province that has a potentially significant effect on SOC content due to large differences in the number of falling leaves and plant residues. The MrVBF shows that flat valley bottoms where sediments and outflows accumulate leading to higher clay and SOC contents [34,67]. All environmental variables which did not conform with SOC grid resolution of 30 × 30 m were resampled to a 30 m spatial resolution using either the nearest neighbor or bilinear resampling methods.

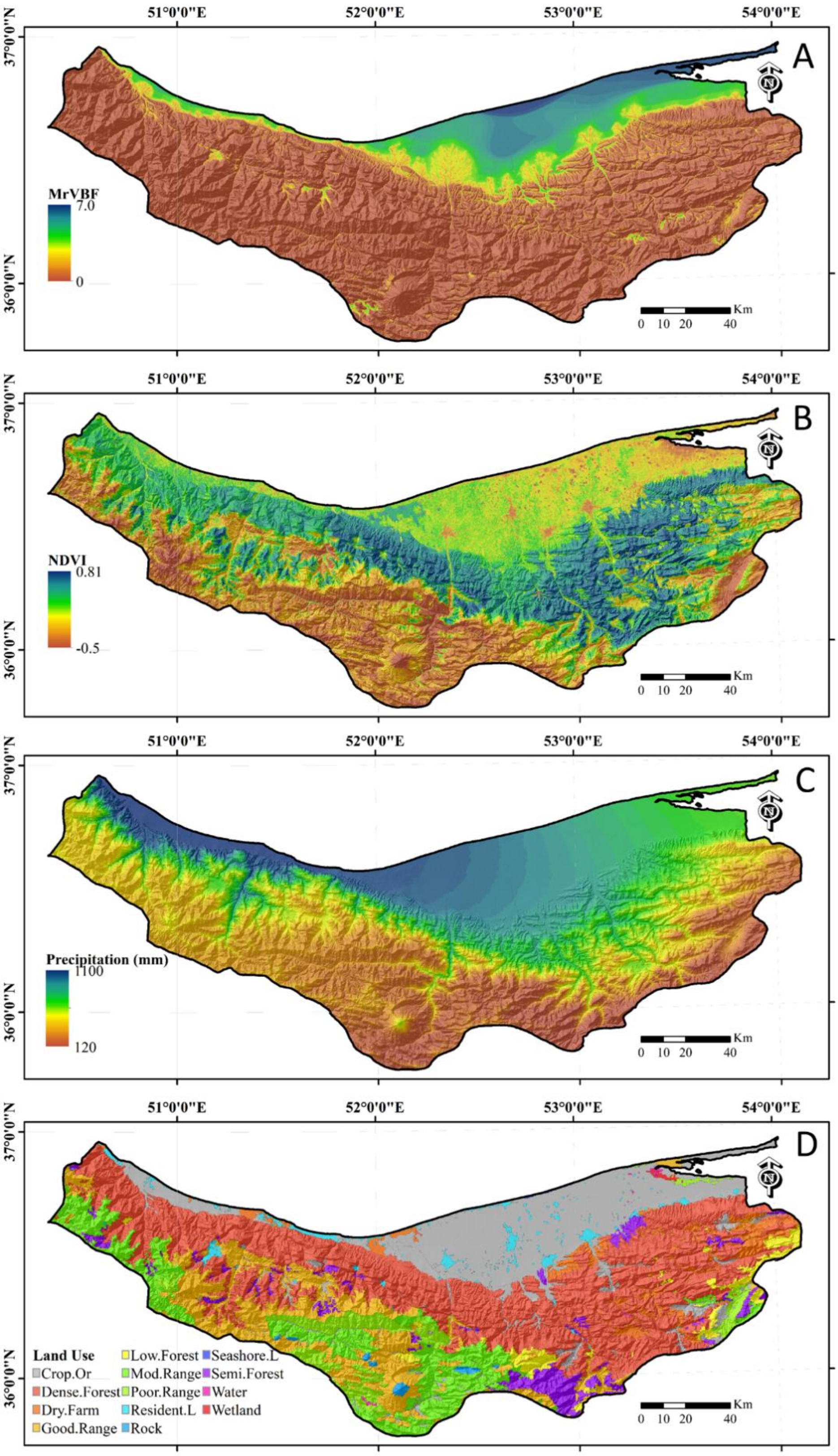

**Figure 2.** Four examples of auxiliary variates used for modeling soil organic carbon: (**A**) multiresolution valley bottom flatness index (MrVBF), (**B**) normalized difference vegetation index (NDVI), (**C**) precipitation, and (**D**) land use maps. The Crop.Or, Dense.Forest, Dry.Farm, Good.Range, Low.Forest, Mod.Range, Poor.Range, Resident.L, Seashore.L, and Semi.Forest symbols in land use map correspond to the land use type of croplands and orchards, dense forestlands, dry farming lands, good rangelands, low dense forestlands, moderate rangelands, poor rangelands, residential lands, seashore lands, and semidense forestlands, respectively.

## 2.4. Selection of Auxiliary Variables Using Genetic Algorithms (GA)

Instead of taking all 105 environmental auxiliary variables into consideration for a predictive ML algorithm, the feature selection method reduces the number and collinearity of the auxiliary variables. The most informative auxiliary variables should be inserted into the algorithms with the aim of high accuracy of the ML algorithms for SOC prediction [9,16,25]. The selection of the significant environment auxiliary variables is a preprocessing step for ML algorithms to remove redundant and irrelevant variables. For this study, one of the most advanced algorithms for feature selection, namely the genetic algorithm (GA), was used to select the most appropriate auxiliary data to be fed as inputs to the ML algorithms [16]. GA is able to select those auxiliary data that are not only essential but improve performance as well. Moreover, GA can manage the nonlinear relationships between SOC and auxiliary data [70].

By mimicking natural biological evolution, the GA which is a heuristic search algorithm provides the best value for a function [51]. A GA feature selection process starts with an initial random population consisting of individuals. The individuals, representing subsets of auxiliary data, are encoded as binary in which 1 represents if the feature is selected and 0 otherwise [71]. Then three primary operations including selection, crossover, mutation repeat until a stopping criterion is reached. The selection operations were for selecting the two fittest individuals for reproduction (i.e., the solutions providing the lowest root mean squared error, RMSE). The crossover recombines two individuals to create new ones which may be better. The mutation operator introduces alteration in a small number of individuals. The process of selection, crossover, and mutation continues until a termination condition is satisfied [48,52]. Importantly, for each generation, it is necessary to assign a fitness value to each individual in the population so that the RMSE values are calculated by fitting the random forest model [46,48,52].

In this study, the GA procedure was performed with 10-fold cross-validation and 100 iterations to select the smallest number of auxiliary variables important for SOC modeling using the caret package in R [72]. The population size, crossover, and mutation rates used were 50, 0.6, and 0.001, respectively, as outlined by Welikala et al. [52].

## 2.5. Machine Learning Techniques

In this study, six ML algorithms including support vector machines (SVM), artificial neural networks (ANNs), regression tree (Cubist), random forest (RF), extreme gradient boosting (XGBoost), and deep neural networks (DNN) were chosen. Each algorithm can discover complex relationships between SOC content and auxiliary covariables. Table 1 summarizes the hyperparameters of the six ML algorithms used in this study. A brief description of the ML techniques used in this study follows.

**Table 1.** Hyperparameters of machine learning algorithms used in this study.

| ML Algorithms | Hyperparameters | Definition | Defined Parameters |
|---|---|---|---|
| SVM (support vector machines) | Kernel type<br>C<br>$\sigma$ | the kernel function<br>the penalty parameter<br>the bandwidth parameter | RBF<br>0.01–100<br>0.01–100 |
| Cubist (regression tree) | committees<br>neighbors | the number of model trees<br>the number of nearest neighbors | 1–100<br>0–9 |
| XGBoost (extreme gradient boosting) | booster<br>max_depth<br>min_child_weight<br>colsample_bytree<br>subsample<br>eta | the type of model<br>the depth of tree<br>the minimum sum of weights of all observations<br>the number of variables supplied to a tree<br>the number of samples supplied to a tree<br>learning rate | gbtree<br>3–10<br>0–5<br>0.5–1<br>0.5–1<br>0.01–0.5 |
| RF (random forest) | Mtry<br>Ntree | the number of input variables<br>the number of trees | 1–30<br>100–3000 |
| ANN (artificial neural networks) | decay<br>size | learning rate<br>the number of neurons in the hidden layer | 0.001–0.05<br>1–10 |
| DNN | hidden | the number of hidden layers | 2–10 |
| (deep neural networks) | size | the number of neurons in the hidden layer | 15–200 |
| | network weight initialization | the initialized weight of networks | uniform/he_normal |
| | learning rate | that controls adjusting the weights of the network | 0.001–0.05 |
| | dropout regularization | the amount of the neurons that are randomly dropped | 0.2–0.8 |

ML: machine learning; SVM: support vector machine; Cubist: regression tree; XGBoost: extreme gradient boosting; RF: random forest; ANN: artificial neural networks; DNN: deep neural networks.

#### 2.5.1. Support Vector Machines (SVMs)

Initially, SVMs were developed as a methodology for resolving problems of classification into two attributes using a threshold value. Connected with the earlier development of SVMs as a classification method, the regressive type of support vector machines was proposed. This caused the spread of the philosophy of support vectors machines being used to solve regression problems. The algorithm was formulated as a linear method and then it was generalized to (1) the presence of noise in the data using slack variables following the soft-margin philosophy, and (2) a nonlinear model through the conversion of the input space into a larger dimension, as done for classification [73]. Hence, SVM is used for classification and regression processes with a set of connected supervised learning algorithms and they have an excellent ability to be universal predictors of any multivariate function to any specified degree of accuracy [20]. In this study, the SVM algorithm was employed by improving the range of its components (C: 0.01–100; $\sigma$: 0.01–100) based on the input data (Table 1).

#### 2.5.2. Artificial Neural Networks

Artificial neural networks (ANNs) stand out among the different types of models because they are calculative techniques with mathematical models simulated from the human's brain neural function [74]. ANNs as vigorous data-modeling tools attain knowledge by way of experience. They are able to detect patterns and draw results, therefore they can be used for data prediction with correlation, such as soil properties. The ability for handling and modeling multiple outputs simultaneously is a primary benefit of ANN techniques [75]. The development of an ANN model mainly consists of three main stages: the generation of data for the training/testing of the model, the selection of the optimal configuration, and the validation of the model on an independent data set. Additionally, ANNs are interconnected by structures called perceptrons and consist of input, output, and hidden layers that transform the input into something that the output layer can utilize [76]. ANN models allow one to attribute lower weight to samples that deviate from a standard, since ANNs can identify patterns in data distribution, which is not observed in linear and nonlinear regression [77]. As a result, ANN models can lead to a higher accuracy than linear and nonlinear regression [78]. The present neural networks of this study were made based on a learning rate of 0.001–0.05 and the number of hidden layer neurons was 2–10 (Table 1). The sigmoid function is the activation function in the nnet package [79] for the MLP with one hidden layer which we used in this study.

#### 2.5.3. Regression Tree (Cubist)

Cubist is an ML estimating tool that is similar in approach to regression trees [80]. However, unlike classification and regression tree models, linear models are often well structured rather than having end values [81]. Additionally, the Cubist and other regression tree algorithms have a clear difference, that is the linear regression model is fitted to the leaf nodes of the trees in the Cubist [82]. Overall, the Cubist approach makes multivariate models that are made of sets of rules, and the prediction model will be chosen based on the rules [83]. In this research, the ptoposed regression tree models are based on Cubist regression models [84]. Cubist models were improved by defining the number of model trees, and nearest neighbors using the data set shown in Table 1.

### 2.5.4. Random Forest (RF)

The random forest (RF) consists of a series of binary rule-based decisions that define relationships between input and its dependent variables. It comprises a large number of individual tree algorithms trained from bootstrap samples of the data [85]. The single prediction will be made by accumulating the results of all trees. One of the main benefits of random forests is that they can precisely explain the compound connections between the independent variables and the dependent variables. So when composite environmental systems and ecological supplementary variables are introduced, RF can be helpful [86]. Two important parameters in RF algorithms are the number of trees (Ntree) and the number of variables (Mtry) which are available for selection in each split [87]. These two main parameters (Mtry and Ntree) were adjusted for the best result. The range of values used is shown in Table 1.

### 2.5.5. Extreme Gradient Boosting (XGBoost)

The algorithm for extreme gradient boosting (XGBoost) was proposed by Chen and Guestrin [88]. It is an algorithm for improving the performance for gradient boosting machines and especially for regression trees and K classification methods [89]. By the supplemental training strategies, the "boosting" as a basic idea of this method extends a "strong" learner from a set of "weak" learners. The XGBoost technique is supposed to improve calculation but also reduce over-estimation events. The XGBoost simplifies the objective functions and improves the calculation speed to an optimum by allowing the combination of estimative and adjustment terms. In addition, in the XGBoost approach, during the training step, simultaneous computations will be done automatically for the functions [89]. More information can be obtained about the XGBoost algorithm from the work of [88]. Table 1 shows the XGBoost algorithm parameters used to do this research including the type of algorithm, the depth of trees, the minimum sum of weights of all observations, the number of variables provided to a tree, the number of samples provided to a tree, and the learning rate.

### 2.5.6. Deep Neural Networks (DNN)

The performance of conventional DNN as an estimation algorithm for remote sensing applications has been extensively explored during the past few years [90]. DNN has been reported as a reliable and efficient approximation function for delivering insight into the relationship (whether a linear or a nonlinear relationship) between input and output variables [91]. DNN has shown promising results in a wide and diverse range of applications from digital signal processing and control systems to hazard susceptibility mapping [92,93]. Figure 3 illustrates the architecture of a conventional DNN. The networks are configured by passing several layers for learning the probability of the outputs.

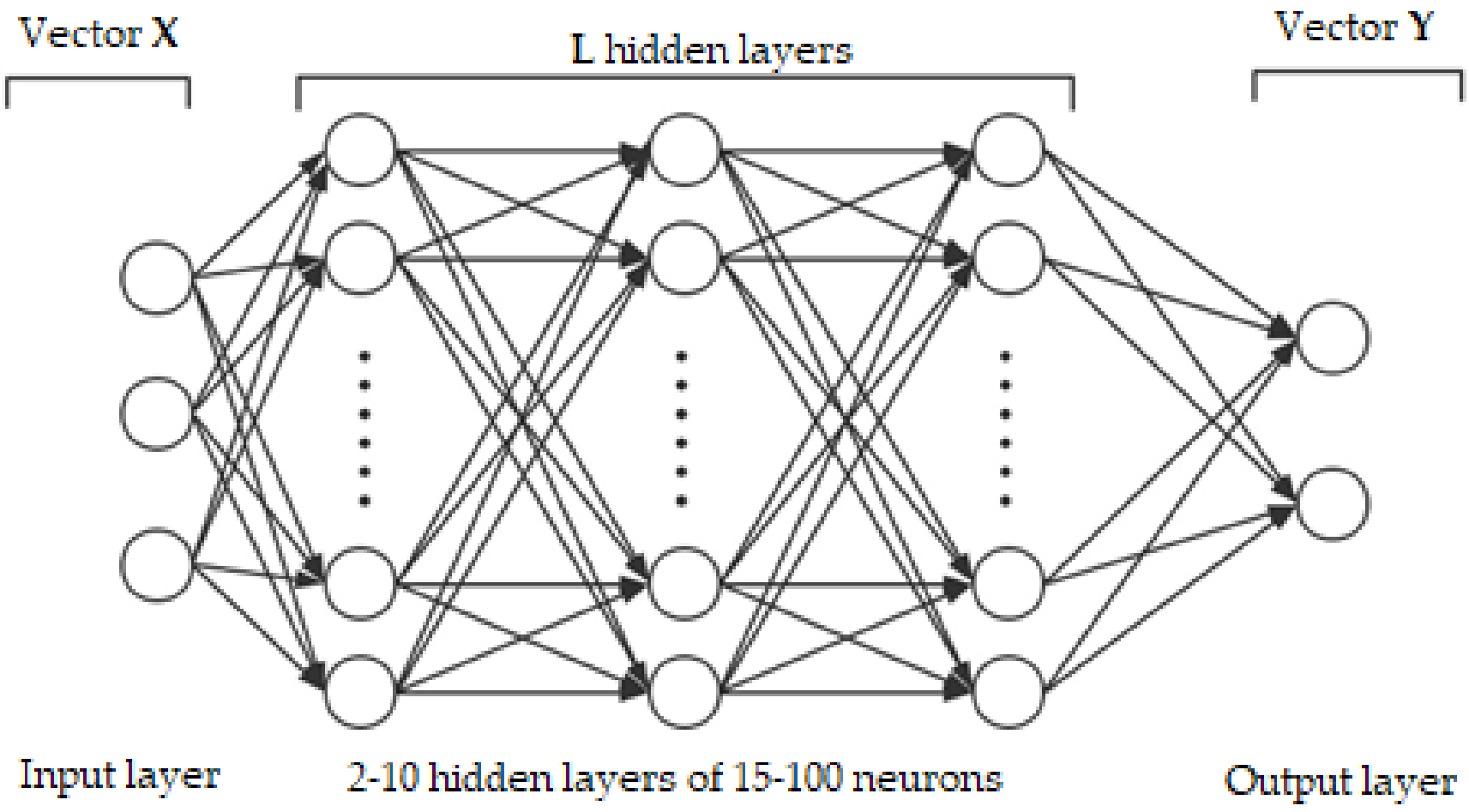

**Figure 3.** Illustration of the architecture of conventional deep neural network (DNN).

In this study, conventional DNN is a feedforward learning network where there is no looping back from the output layer to input. In this case, the DNN produces a map of virtual neurons and random weights. The inputs and weights will be multiplied and would deliver outputs within the range of 0 and 1. The algorithm would adjust the weights to accurately identify a particular learning pattern to fully process the data. The DNN includes L hidden layers, the input layer (vector **X**), and the output layer (vector **Y**). As recently formulated by Wang et al., [94], the estimation of **Y** can be presented as follows.

$$\mathbf{z}^1 = \sigma_1(\mathbf{W}^1\mathbf{X} + \mathrm{b}^1) \tag{1}$$

$$\mathbf{z}^2 = \sigma_2(\mathbf{W}^2\mathbf{z}^1 + \mathrm{b}^2),\quad \mathbf{z}^L = \sigma_L(\mathbf{W}^L\mathbf{z}^{L-1} + \mathrm{b}^L),\quad \mathbf{Y} = \mathbf{W}^{L+1}\mathbf{z}^L + \mathrm{b}^L,\ \ \theta = \{\mathbf{W}^i, \mathrm{b}^i\}_{i=1}^{L+1} \tag{2}$$

where $\mathrm{b}^i$ and $\sigma_i$ are the bias and the activation function of the *i*th layer. Here, $\mathbf{W}^i$ represents the weights. In Equation (2), the $L + 1$ represents the output layer. Therefore, **Y** can be presented as follows.

$$\mathbf{Y} = NN(\mathbf{X}; \theta) \tag{3}$$

Eventually, through calculating the mean square error (MSE) of output and input values, loss function **L** can be estimated as follows.

$$MSE_{Data} = L(\theta) = \frac{1}{N}\sum_{i=1}^{N} |NN(\mathbf{X}_i; \theta) - \mathbf{Y}_i|^2 \tag{4}$$

where *N* represents the sequence of data. Here, the GA is used to minimize the $L(\theta)$ function for the training. A trained DNN is further used for the estimation of the new variables. The predictive ability of neural networks is possible by learning large amounts of data. Generally, input data create the training datasets, and similar output data will be entered into a neural network algorithm. This algorithm can detect the basic rules in the data entered and compose an interior model that is suitable to estimate the new input data using several training repetitions during the process. The model can be computed by the interactions and connections between neurons, whereas any physical or clear mathematical relationships cannot be supplied [90]. The neural network structure can affect the precision of the predictive models. Each latent layer of the DNN algorithm consists of some calculative neurons that are interconnected to the next calculative neurons in the adjoining latent layers. To finalize the DNN model, the neurons of each latent layer measure the calculative neuron outputs of the prior layer, and after the computation procedure of the activation function, the outputs are generated for the subsequent layer [42].

Table 1 shows the specifications used for DNN, which are hidden layers, size, network weight initialization, learning rate, dropout regularization. In this study, for the DNN method, the $H_2O$ package [92] with the rectifier function as a nonlinear transformation was used for DNNs in this study [95]. It is worth mentioning that adhering to a balanced ratio of training and testing is of utmost importance in modeling with machine learning [96]. Several methods in a wide range of applications are introduced to identify the correct balance for testing [97]. Nevertheless, the evaluation metrics have been shown to be reliable measures to maintain a sufficient number of elements for a training dataset in soil research [37–40]. It is often observed that by decreasing the amount of training data, the error increases, which accurately indicates the worth of data for models. The amount of training data, in this study, is optimally tuned to ensure the lowest errors. The total dataset is divided into 10 datasets that are sequentially used for training and testing. The DNN is calibrated 10 times to assure each data point was used as validation at least once.

### 2.6. Evaluation of Algorithm Performance

Ten-fold cross-validation was implemented for testing the performances of six ML prediction algorithms for estimating the SOC contents in Mazandaran province. In this regard, the total dataset was split into 10 datasets that were sequentially used as training and testing datasets for a given prediction algorithm. Each prediction model is calibrated 10 times, guaranteeing each data point was used as validation at least once. Then, the 10 prediction errors can be obtained for each prediction algorithm. The four evaluation criteria used in this study are the coefficient of determination ($R^2$) [98], Lin's concordance correlation coefficient (CCC) [99], mean absolute error (MAE) [100] and root mean squared error (RMSE) [100] with following formulas:

$$R^2 = 1 - \left(\frac{\sum_{i=1}^{n}(Oi - p)^2}{\sum_{i=1}^{n}(Oi - O')^2}\right) \quad (5)$$

$$CCC = \frac{2\, r\, \sigma_o \sigma_p}{\sigma_o^2 + \sigma_p^2 + [O' - P']^2} \quad (6)$$

$$MAE = \frac{1}{n}\sum_{i=1}^{n}|Pi - Oi| \quad (7)$$

$$RMSE = \sqrt{\frac{1}{n}\sum_{i=1}^{n}(Pi - Oi)^2} \quad (8)$$

where, $n$ is the number of samples, $Oi$ and $Pi$ are observed and predicted SOC contents, respectively. $O'$ and $P'$ are the means for the observed and predicted SOC contents, respectively. Furthermore, $\sigma_o$ and $\sigma_p$ are the variances of observed and predicted values. Four criteria of the validation datasets in 10-fold validation in each prediction algorithm were averaged and used for selecting the best performing prediction algorithms. The prediction algorithm with the lowest MAE and RMSE, and highest $R^2$ and CCC values are determined as the best for SOC prediction.

### *2.7. Uncertainty Assessment*

The spatially explicit quantification of the uncertainty of SOC prediction is analyzed in this study. The SOC maps generated by each model were used to calculate the mean and standard deviation (SD) of the SOC for each pixel in 10-fold realization [101]. It was assumed that six ML models follow the normal distribution for each raster cell. The confidence interval (CI) was calculated with the mean as ±1.64 SD for a given 90% CI. The upper and lower limit of the 90% CI were mapped. The mean of the SOC contents in each pixel and the 90% CI was calculated by retrieving the 5th and 95th percentiles of prediction. Finally, three maps of the SOC were produced for the best performing model: the mean prediction, lower CI (5%), and higher CI (95%).

## 3. Results and Discussions

### *3.1. Summary Statistics*

Table 2 shows the summary statistics for topsoil SOC for the 1879 sampling sites. SOC contents ranged from 0.02% to 11.48% with a mean and standard deviation of 2.19% and 1.27%, respectively. The lower SOC contents correspond to highly degraded lands where the surface soil is eroded and the maximum SOC contents were observed in dense forestlands. The coefficient of variation of 58.23% demonstrates the high variability of SOC contents within the study area. The values of skewness (2.33) and kurtosis (8.2) indicate that the SOC data is highly skewed and in turn, violates assumptions of normality. SOC data were anchored at 1.00 and then, transformed by the natural logarithm to make the distribution less skewed. The skewness and kurtosis values of the log-transformed SOC values were 0.59 and 1.50, respectively, and the Kolmogorov–Smirnov test showed that the distribution of these log-transformed values was not significantly different from normal. Further analysis was performed on the log-transformed data; and the predicted SOC values were back-transformed to the original scale.

**Table 2.** Descriptive statistics of SOC (%) in this study ($n$ = 1879).

| Min | Max | Mean | SD | CV | Skewness | Kurtosis |
|---|---|---|---|---|---|---|
| 0.02 | 11.48 | 2.19 | 1.27 | 58.23 | 2.33 | 8.2 |

SOC: soil organic carbon; Min: minimum; Max: maximum; SD: standard deviation; CV: coefficient of variation; K-S $p$-value: significance level of Kolmogorov–Smirnov test.

### *3.2. Selected Auxiliary Data*

In this study, the GA procedure with 10-fold cross-validation and 100 iterations (i.e., GA was executed 1000 times) was used to select the minimum number of important auxiliary variables for SOC modeling. The results of GA for 1000 generations are presented in Figure 4. The average of the internal out-of-bag RMSE estimates as well as the average of the external performance estimates calculated. Based on these results, the generation associated with the best external RMSE estimate was 0.86. The GA selected 35 predictors out of 105 environmental variables, as the most relevant driving factors for SOC mapping in Mazandaran province (Table 3). These variables comprised 13 terrain attributes, 18 remotely sensed variables, two climatic variables, and two categorical data layers. The resolution and origins of the 35 predictors are given in Table 3. Getting more information about the contribution of each variable to SOC variability is of great importance. Therefore, the significance of each environmental auxiliary variable was analyzed using a sensitivity analysis and was represented as an attribute percentage. Figure 5 indicates the order of the relative importance of the selected predictors on SOC spatial variability using the procedure outlined by [45].

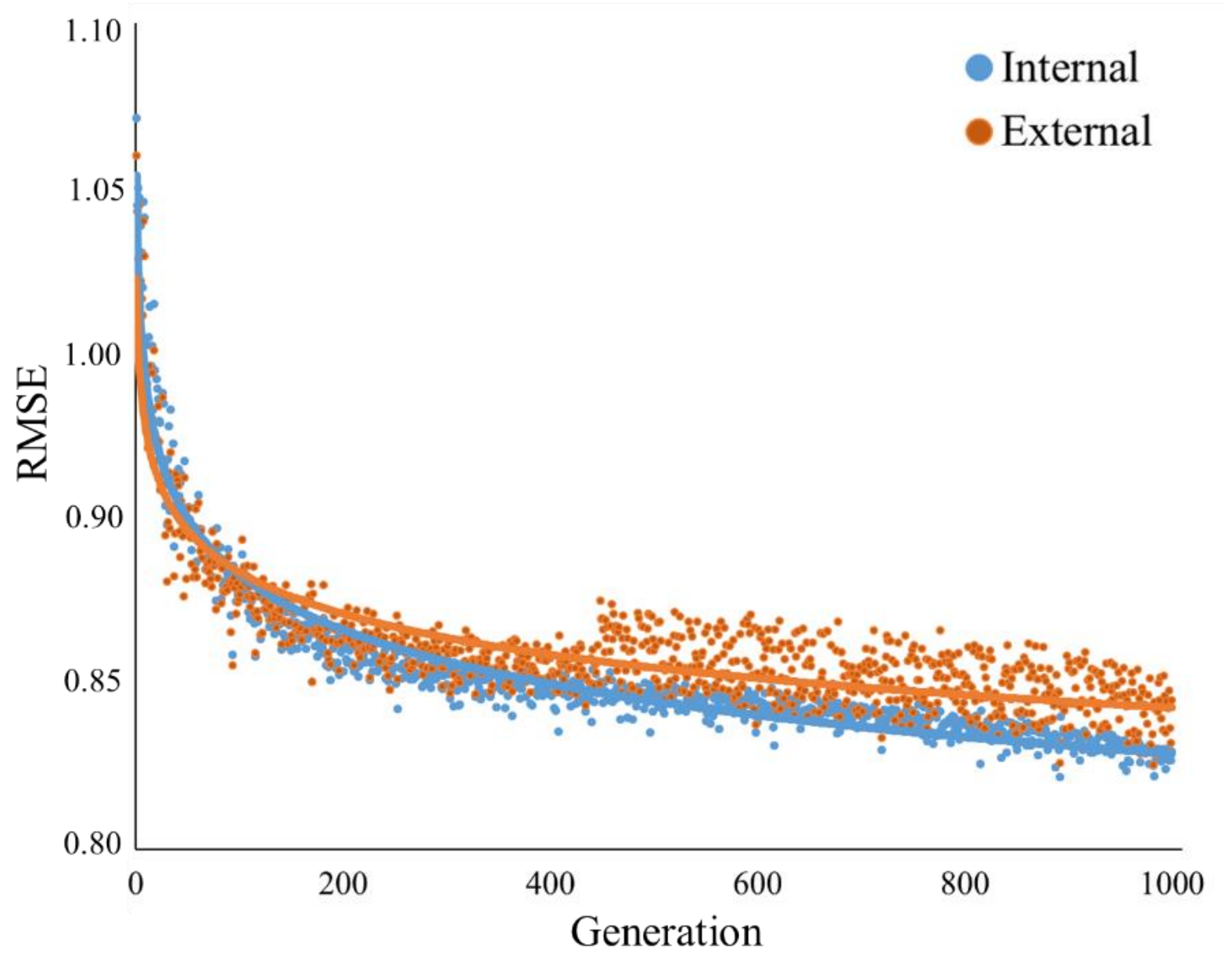

**Figure 4.** Internal and external validations of GA.

**Table 3.** The selected auxiliary data using genetic algorithms for predicting SOC.

| Definition | Res. | Ref. |
|---|---|---|
| ***Selected Terrain Attributes*** | | |
| Aspect | 30 m | SRTM |
| Slope Gradient | 30 m | SAGA GIS |
| Elevation | 30 m | SAGA GIS |
| Slope Length Factor | 30 m | SAGA GIS |
| Valley Depth | 30 m | SAGA GIS |
| Openness (PosOpen) | 30 m | SAGA GIS |
| Openness (NegOpen) | 30 m | SAGA GIS |
| Catchment Slope (CaSLOP) | 30 m | SAGA GIS |
| Plane Curvature (Plan.Curv) | 30 m | SAGA GIS |
| Topographic Wetness Index (TWI) | 30 m | SAGA GIS |
| Channel networks base level (CHNL.BASE) | 30 m | SAGA GIS |
| Multiresolution ridge top flatness index (MRRTF) | 30 m | Gallant and Dowling (2003) |
| Multiresolution Valley Bottom Flatness Index (MrVBF) | 30 m | |
| ***Selected RS data*** | | |
| Blue band of Landsat-8 (B1) | 30 m | Wulder et al. (2016) |
| Green band of Landsat-8 (B2) | 30 m | |
| Red band of Landsat-8 (B3) | 30 m | |
| Near-infrared band of Landsat-8 (B4) | 30 m | |
| Shortwave IR-1 band of Landsat-8 (B5) | 30 m | |
| Shortwave IR-2 band of Landsat-8 (B6) | 30 m | |
| Normalized difference vegetation index (NDVI) | 30 m | Rouse et al. (1974) |
| Enhanced vegetation index (EVI) | 30 m | |
| Combined Spectral Response Index (COSRI) | 30 m | |
| Transformed SAVI (TSAVI) | 30 m | |
| Soil adjusted vegetation index (SAVI) | 30 m | |
| Brightness Index | 30 m | Metternicht and Zinck (2003) |
| Clay Index | 30 m | Boettinger et al. (2008) |
| Carbonate index | | |
| MODIS Red | 250 m | |
| MODIS Near Infrared (MODIS Nir) | 250 m | |
| MODIS Night Temperature(MODIS.Night.Temp) | 1000 m | |
| MODIS Day Temperature (MODIS.Day.Temp) | 1000 m | |
| ***Selected climatic data*** | | |
| Annual precipitation (mm) | 1000 m | Fick and Hijmans (2017) |
| Annual mean temperature (°C) | 1000 m | Fick and Hijmans (2017) |
| ***Selected categorical data*** | | |
| Land use | 125 m | Banaei et al. (2005) |
| Soil map | 500 m | |

Abr.: abbreviation; Res.: resolution; Ref: references.

As can be seen in Figure 5, precipitation is the most crucial feature (14.9%) driving the spatial variability of SOC contents in Mazandaran province followed by the NDVI (12.5%), MODIS day temperature (10.6%), MrVBF (8.7%), land use (8.2%), valley depth (7.2%), and MODIS night temperature, respectively. In Mazandaran province, as previously thought, the precipitation significantly affects SOC contents by enhancing vegetation coverage and the rate of organic matter inputs. Together precipitation and temperature (MAT) explain 18.9% of the variation in SOC contents demonstrating the high dependencies of SOC contents in the province-scale to the climatic variables. Lamichhane et al. [14] reviewed several studies and pointed out that climate is the most influential factor for the variation of SOC at large extents. The high precipitation is mostly combined with lower temperatures and slower SOC decomposition rates at higher altitudes [102,103]. Falahatkar et al. [104] reported that the most important auxiliary predictors for SOC stocks for surface soil in Guilan province, northern Iran are land use, NDWI, silt, clay, and elevation. Along with our findings, the surface temperature data derived from remotely sensed data (Landsat) was found to be influential for improving SOC prediction [66].

The NDVI is the second most important feature explaining SOC variability, indicating that the SOC contents were highly influenced by vegetation variation. SOC content was highly dependent on the natural vegetation cover intensity, and the plant residue left after plant harvesting [4,36,105]. Due to the dependency of SOC on vegetation cover, NDVI has frequently been used as a predictor for mapping SOC in several studies [16,25,33,106]. Additionally, NDVI has more importance to SOC contents compared to other remotely sensed vegetation indices like EVI. Although the EVI performs better than the NDVI in many applications, our results indicated that NDVI is more important for explaining SOC contents compared to EVI. It might be related to the topographic conditions as Matsushita et al. [107] reported that EVI is more sensitive to topographic conditions than is the NDVI. Meanwhile, the green band of Landsat-8 (B2) showed a higher contribution to SOC than B3 and B4.

The high contribution of the MrVBF (8.7%) to SOC variability in this study could be attributed to the deposition of fine organic-enriched particles and sediment [58] from the highlands in the lower valleys with flat and low-lying areas. Land use data is considered as the fifth important variable for SOC variation. Land use effects on SOC variation is related to the land-use conversion in the last two decades in Mazandaran province that have led to the exposure of soils and the rapid decomposition of SOC [58].

The TWI only contributed 3% SOC contents showing that SOC tends to accumulate in wetter, low-lying areas in Mazandaran province. Taghizadeh-Mehrjardi et al. [25] demonstrated that the wetness index is the most important terrain variable for the prediction of SOC in subsoils (>30 cm depth). Plan curvature (Plan.Curv), soil map, slope, aspect, channel networks base level (CHNL.BASE), LS factor and other variables ranked in Figure 4 made only a small contribution to SOC spatial variability at the province-scale used in this study but still need to be considered as input variables for SOC modeling by ML techniques.

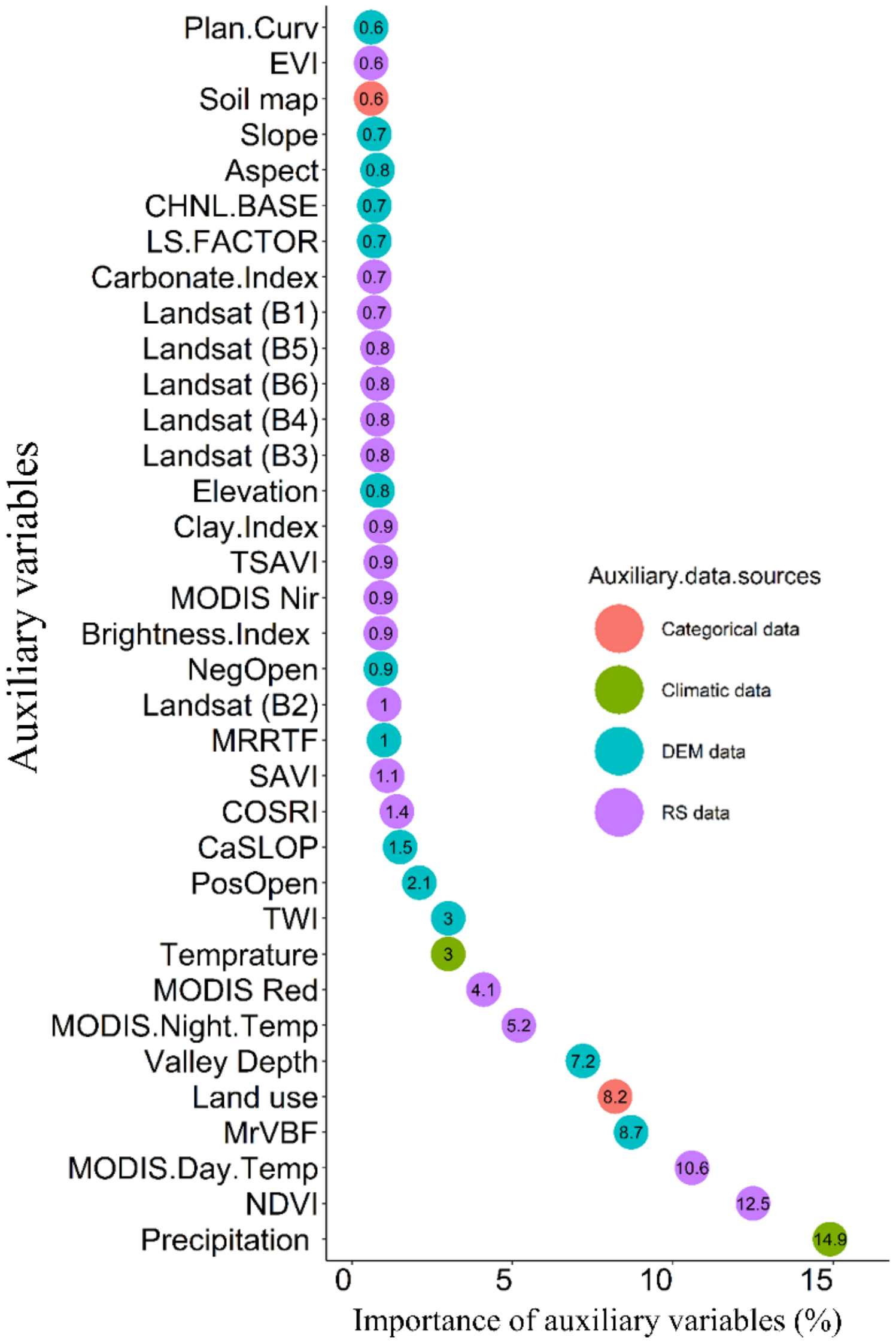

**Figure 5.** Relative importance of auxiliary data using genetic algorithms. (Refer to Table 3 for a definition of auxiliary data).

## 3.3. Machine Learning Performances

The average MAE, RMSE, $R^2$, and CCC for SOC prediction by 10-fold cross-validation are shown in Table 4. The proposed ML models showed different abilities to predict SOC contents at unsampled locations at the province-scale. This could be related to the various mathematical functions of each algorithm [2]. The mean $R^2$ values indicate that the SVM, ANN, Cubist, RF, and XGB models deliver 53%, 55%, 57%, 58%, and 57% of SOC variability, respectively. However, the DNN model outperforms other models by delivering 65% of the SOC variability. In all ML models, the RMSE values are reported more significant than the MAE, indicating that there is a contribution of the errors in SOC predictions [108]. The DNN algorithm showed the lowest mean MAE value (0.59%) of the six studied ML algorithms. The SVM algorithm had the highest error with mean RMSE values of 0.87% compared with other ML models, meanwhile, the DNN outperformed with the lowest mean RMSE value (0.75%). ANN, Cubist, RF, and XGB showed a similar ability to predict SOC in Mazandaran province. Based on the performance criteria used, SVM was always a weaker ML algorithm than the other algorithms, while DNN was the most consistently robust algorithm.

One of the main advantages of DNN is that the step of feature extraction was performed by the DNN model itself [10]. Our results confirm previous research on the performance of DNN in soil modeling. For instance, a recent research study introduces the DNN as an effective and robust modeling method to capture the complex nonlinearity between auxiliary variables and soil moisture [40] and SOC prediction [11,42]. DNN models by using the multiple hidden layers of the neural network improve the SOC prediction. Padarian et al. [11] reported that using deep learning models for digital soil mapping offers a simple and effective framework for future soil mapping. The DNN algorithm needs a large number of parameters to be fitted so that it performs well with a large dataset like the one used in this study. Importantly, the sample size is a critical issue for training in the DNN [41].

**Table 4.** Comparisons of the accuracy of six machine learning models for validation dataset by 10-fold cross-validation (means ± standard deviation).

| ML Algorithms | MAE | RMSE | $R^2$ | CCC |
|---|---|---|---|---|
| SVM | 0.69 ± 0.07 | 0.87 ± 0.05 | 0.53 ± 0.05 | 0.76 ± 0.05 |
| ANN | 0.67 ± 0.08 | 0.85 ± 0.07 | 0.55 ± 0.05 | 0.77 ± 0.06 |
| Cubist | 0.66 ± 0.06 | 0.83 ± 0.04 | 0.57 ± 0.04 | 0.78 ± 0.04 |
| RF | 0.65 ± 0.03 | 0.82 ± 0.03 | 0.58 ± 0.05 | 0.78 ± 0.03 |
| XGB | 0.66 ± 0.04 | 0.83 ± 0.04 | 0.57 ± 0.03 | 0.78 ± 0.04 |
| DNN | 0.59 ± 0.06 | 0.75 ± 0.06 | 0.65 ± 0.05 | 0.83 ± 0.06 |

ML: machine learning; SVM: support vector machine; Cubist: regression tree; XGBoost: an extreme gradient boosting; RF: random forest; ANN: artificial neural networks; DNN: deep neural networks; MAE: mean absolute error; RMSE: root mean square error; $R^2$: the coefficient of determination; CCC: Lin's concordance correlation coefficient.

underestimated. The CCC statistic quantified the level of agreement between predicted and measured SOC values according to the 1:1 line. It is based on CCC values (Table 4). The The lower performance of other ML algorithms except for the DNN could be related to taking a large number of auxiliary variables into account and the original data having multiple-scales of variation, as well as different sources and sampling times, all of which increase the uncertainty.

The 1:1 scatterplots of actual vs. predicted SOC using the six ML algorithms are shown in Figure 6. It is now much easier to understand the prediction efficiency of the DNN algorithm as most predictions follow the 1:1 line with the exception of large observed SOC contents, which were slightly DNN algorithms with a 0.83 value were superior and the SVM (CCC = 0.76) was inferior.

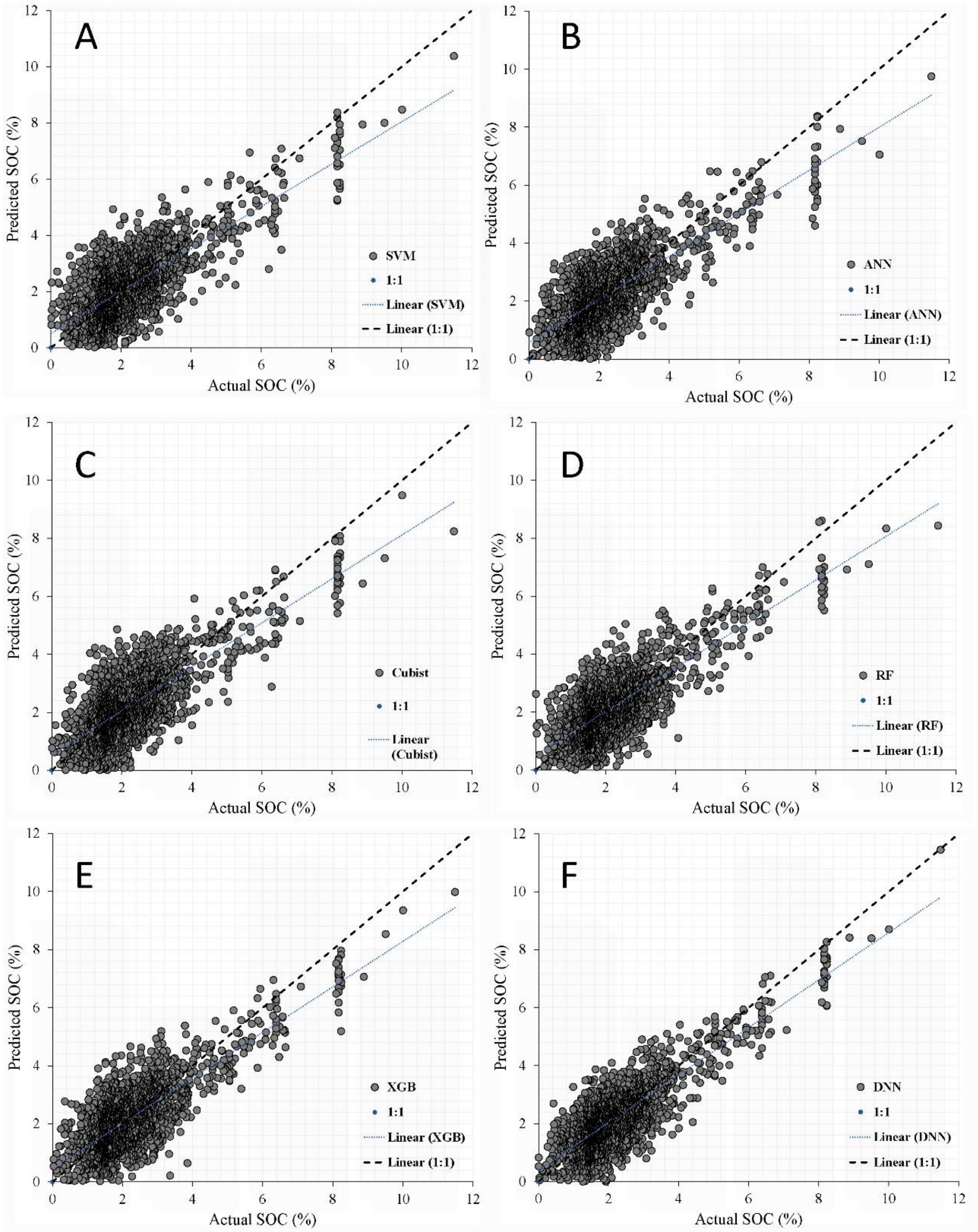

**Figure 6.** Actual vs. predicted values of soil organic carbon using six machine learning algorithms: (**A**) SVM, (**B**) ANN, (**C**) Cubist, (**D**) RF, (**E**) XGB, and (**F**) DNN. (SOC: soil organic carbon; SVM: support vector machine; Cubist: regression tree; XGBoost: extreme gradient boosting; RF: random forest; ANN: artificial neural networks; DNN: deep neural networks).

As the deep learning method is sensitive to the size of the training dataset, DNN apparently yielded the best result in this study due to the large data for training. Using a deep convolutional neural network trained with a smaller dataset was not effective for the prediction of soil properties by spectral data [41].

The DNN algorithm has a more flexible structure and is explicitly able to extract more information from the environmental auxiliary variables and the SOC content and this is consistent with the results of [11,42]. Therefore, we could recommend the DNN algorithm as the best ML algorithm for the prediction of SOC content with reliable uncertainty in Mazandaran province. The DNN model in this study was successfully trained with a large number of covariates due to its more flexible neural structure and in turn was able to effectively combine with multiscale properties. This algorithm uses the data imputation for taking the missing values into account thereby the better model performance could be achievable especially for subsurface DSM [11].

The observed mean value of $R^2$ and RMSE in the DNN algorithm can be compared to the other studies at the regional scale [109–111]. Wang et al. [16] only achieved an $R^2$ mean value of 48% of the total spatial SOC variability using the RF algorithm in semiarid pastures of eastern Australia.

### *3.4. Spatial Prediction of SOC with Uncertainty Estimates*

The numbers and percentages of SOC contents that fall within the 90% CI are shown in Table 5. The uncertainty analysis also showed to some extent the same trend to the ability of the ML algorithms to predict SOC. The DNN had the maximum percentage of observations (~88%) that fell within the defined CI. The spatial prediction of mean SOC content with a 5% lower confidence limit and 95% upper confidence limit values in Mazandaran province produced by the DNN algorithm are shown in Figure 7. It is clear that the combined influence of selected auxiliary variables controls the SOC contents. It is evident that the SOC contents tend to be higher in a strip from the west (more than 3%) to the east (lower than 1%) in the middle of the Mazandaran province. The precipitation gradient and NDVI index, which were the most highly correlated variables, were greatly responsible for SOC variations. The SOC content coincided in a systematic way with increasing the precipitation gradient, NDVI, and MrVBF indices (Figure 2). It was clear from the predicted SOC map that the amounts were higher in the area with high NDVI values ranging from 0.71 to 0.81 (the central part of the study area). The higher rainfall favors higher net primary production of plant residues and explains the higher SOC contents in the middle portion of the province. There is some uncertainty in the predicted map that may be related to the high variability in SOC data, low precision of predictors, inherently poor relationships between SOC and auxiliary variables, and errors in modeling [112]. Considering the multiple scales of auxiliary variables and the good resolution of soil erosion/deposition data, such data can potentially [32] reduce the spatial prediction uncertainty in future studies.

The SOC contents change over time, thus the predicted map can be used as a base-line to indicate temporal changes. Together with the estimation of uncertainty, the prepared maps are more reliable and could be useful for future SOC inventories and province-scale accounting and carbon balance studies.

**Table 5.** The numbers and percentages (%) of SOC that fall within the 90% prediction intervals predicted by machine learning models for 10-fold cross-validation.

| | | Number of Points | | | % | | |
|---|---|---|---|---|---|---|---|
| **ML Models** | **All** | **Inside CI** | **Outside CI** | | **Inside CI** | **Outside CI** | |
| | | **5 to 95%** | **<5%** | **>95%** | **5 to 95%** | **<5%** | **>95%** |
| SVM | 1879 | 1490 | 187 | 202 | 79.30 | 9.95 | 10.75 |
| ANN | 1879 | 1524 | 165 | 190 | 81.11 | 8.78 | 10.11 |
| Cubist | 1879 | 1580 | 155 | 144 | 84.09 | 8.25 | 7.66 |
| RF | 1879 | 1559 | 150 | 170 | 82.97 | 7.98 | 9.05 |
| XGB | 1879 | 1587 | 140 | 152 | 84.46 | 7.45 | 8.09 |
| DNN | 1879 | 1650 | 110 | 119 | 87.81 | 5.85 | 6.33 |

ML: machine learning; SVM: support vector machine; Cubist: regression tree; XGBoost: extreme gradient boosting; RF: random forest; ANN: artificial neural networks; DNN: deep neural networks; CI: confidence interval.

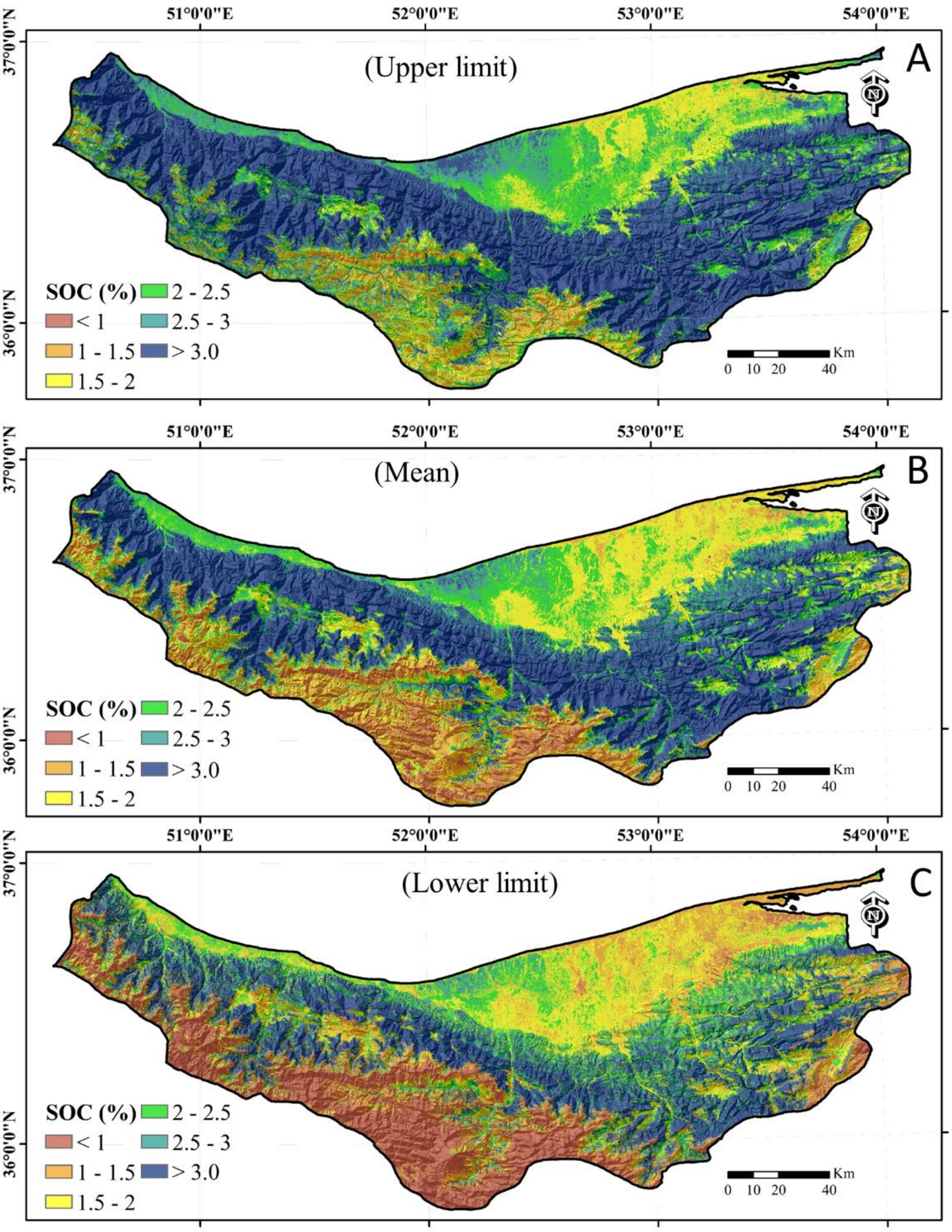

**Figure 7.** Spatial prediction of (**A**) upper, (**B**) mean, and (**C**) lower confidence limits of soil organic carbon (SOC) using a deep neural network model.

### *3.5. SOC Contents in Soil Classes and Geological Eras*

The mean comparison of the SOC contents within different soil orders and suborders is shown in Table 6. The Ultisols and Mollisols with mean SOC contents of 4.04% and 3.20%, respectively, had higher surface SOC compared with other soils. Most Ultisols and Mollisols were found in the dense forest in Mazandaran province showing the higher C inputs into the soils. The high precipitation with a relatively low MAT in the center of the Mazandaran province leads to higher SOC accumulation at the soil surface and in turn higher clay content in a Bt horizon [6,113] and the deeper Bk horizon [6,113,114]. Entisols had the highest SOC variability (CV = 40.66%) followed by the Inceptisols (33.94%), Alfisols (CV = 33.64%), and Mollisols (30.55%). The SOC under Mumults had the highest SOC with the lowest SOC variability (CV = 15.63) whereas Fluvents had significantly the lowest SOC due to the higher SOC decomposition caused by the exposure (tillage) and loss of SOC by erosion.

**Table 6.** The SOC changes in different soil orders and suborders in Mazandaran province.

| Soil Orders | Mean [a] | CV (%) | Soil Suborders | Mean [a] | CV (%) |
|---|---|---|---|---|---|
| Inceptisols | 2.45 B | 33.94 | Aquept | 2.85 C | 20.11 |
| | | | Xerepts | 2.06 B | 36.26 |
| Alfisols | 2.55 B | 33.64 | Aqualfs | 1.94 C | 21.17 |
| | | | Udalfs | 3.17 B | 27.44 |
| Entisols | 2.78 AB | 40.66 | Aquents | 2.51 BC | 14.91 |
| | | | Fluvents | 1.91 C | 22.81 |
| | | | Orthents | 3.93 A | 23.52 |
| Mollisols | 3.20 A | 30.55 | Aquolls | 2.43 BC | 21.64 |
| | | | Rendols | 4.03 A | 22.43 |
| | | | Udolls | 3.33 B | 24.08 |
| | | | Xerolls | 2.31 BC | 34.97 |
| Ultisols | 4.04 A | 12.63 | Humults | 4.04 A | 15.63 |

[a]: Values with different letters in each column indicate significant differences ($p < 0.05$); CV: coefficient of variations.

The mean SOC contents spread across the Mazandaran province differed in soils under different soil SMR and STR classes as shown in Figure 8. The SOC mean value was the highest in the udic SMR class with mean values of 3.85% followed by the aquic (2.45%) and xeric (2.10%), respectively. The high precipitation for soils with the udic SMR class [57] led to the high aboveground biomass production inputs. The greater SOC contents in soils having the aquic SMR class compared to the xeric SMR could be related to anaerobic (reducing) conditions decreasing the rates of organic matter decomposition [115].

The soils having mesic STR classes have high SOC content with a mean value of 2.75%, which was significantly higher than the thermic (2.20%) and cryic (1.25%) STR classes, respectively (Figure 8). Soils in the thermic STR class with higher MAT in low-lying areas had lower SOC contents compared with mesic STR classes reflecting a negative effect of MAT on SOC contents in Mazandaran province due to the high SOC decomposition rate. The small area of the province with a cryic STR class has low vegetation cover in high-altitude lands accompanied by unsuitable temperature conditions for plant growth leading to low inputs of plant residues and biomass. Overall, SOC has been increased with precipitation and decreased with temperature associated with a given altitude in the study area. Moreover, soils formed on younger geological formation have lower SOC contents. The mean SOC contents in soils under the Cenozoic geological era (2.35%) had significantly lower SOC contents compared with soils under Mesozoic (3.12%), Paleozoic (3.35%) and Proterozoic (3.29%) eras, respectively. The higher observed SOC developed on the older geological formations could be attributed to the increased time for SOC to develop and aboveground carbon inputs by the dense vegetation cover inducing the SOC accumulation.

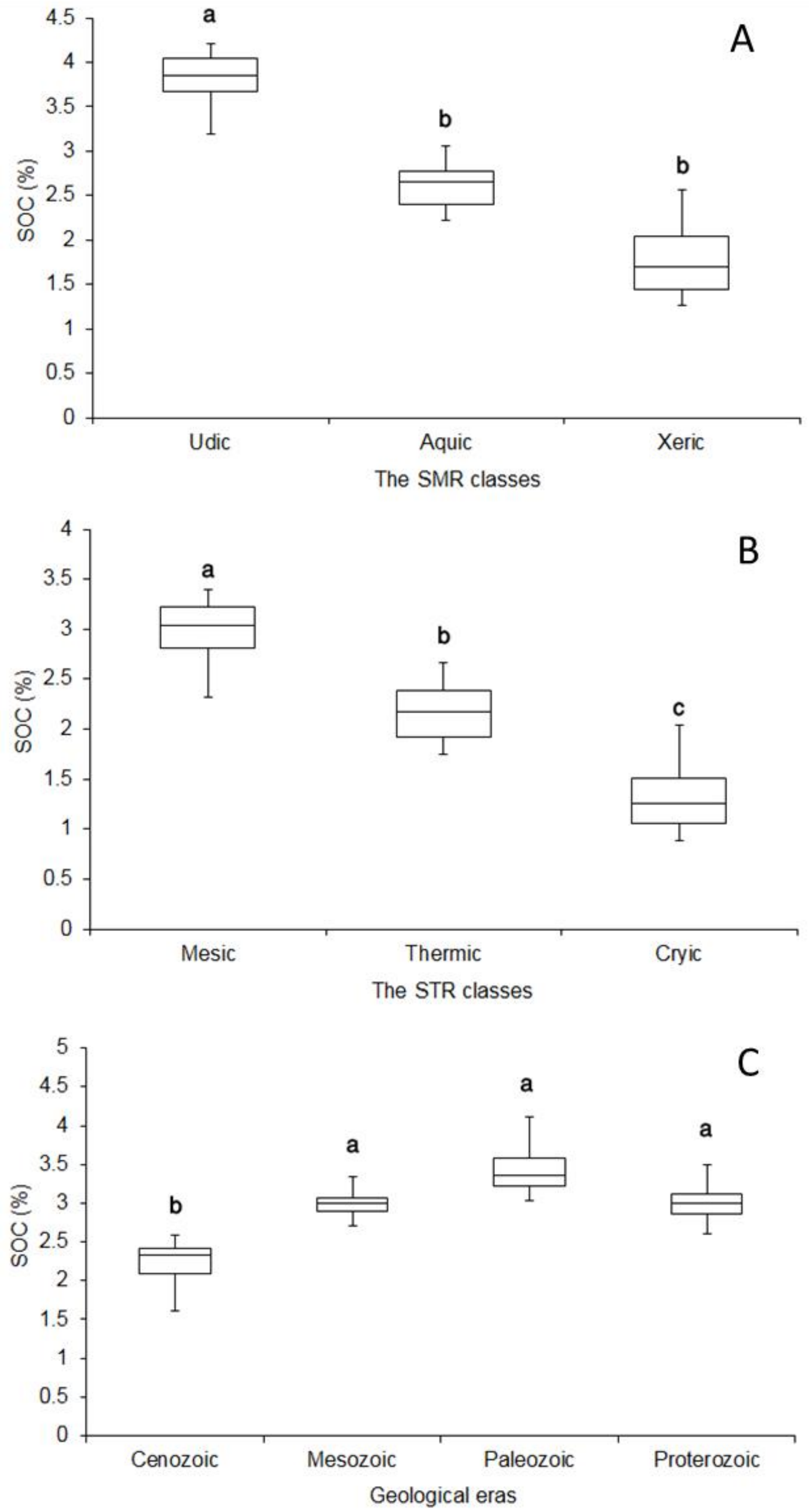

**Figure 8.** Mean comparison of soil organic carbon values within (**A**) soil moisture regime, soil moisture regime (SMR), classes, (**B**) the soil temperature regime, soil temperature regime (STR), classes, and geological eras in the Mazandaran province. Values with different letters in each column indicate significant differences ($p < 0.05$). The maps of SMR and STR classes were produced by Emadi et al. [57] according to the Newhall model.

### 3.6. SOC Contents in Landform Units and Land Uses

The soils formed on mountainous landforms had the highest SOC (3.11%) values with forest land use, while there was little difference in SOC contents for the other landforms except for the soils developed on alluvial fans that had significantly the lowest SOC contents (1.57%) with high coarse fractions (soil particles greater than 2 mm) (Figure 9). The alluvial fans with unstable landforms, have a high susceptibility to erosion and have little water holding capacity providing the lowest aboveground biomass production in the study area. The high degrees of stability in mountain landforms especially in the summit areas [6] showed more developed soils including the Ultisols, Mollisols, and Alfisols. These are closer to the steady-state conditions relative to the younger landforms (Fluvents) leading to greater humification and SOC accumulation on mountain landforms that are currently covered by dense forest. On more geomorphically dynamic/unstable landforms, organic layers can be removed from the developing surface inducing the SOC losses through erosion [115].

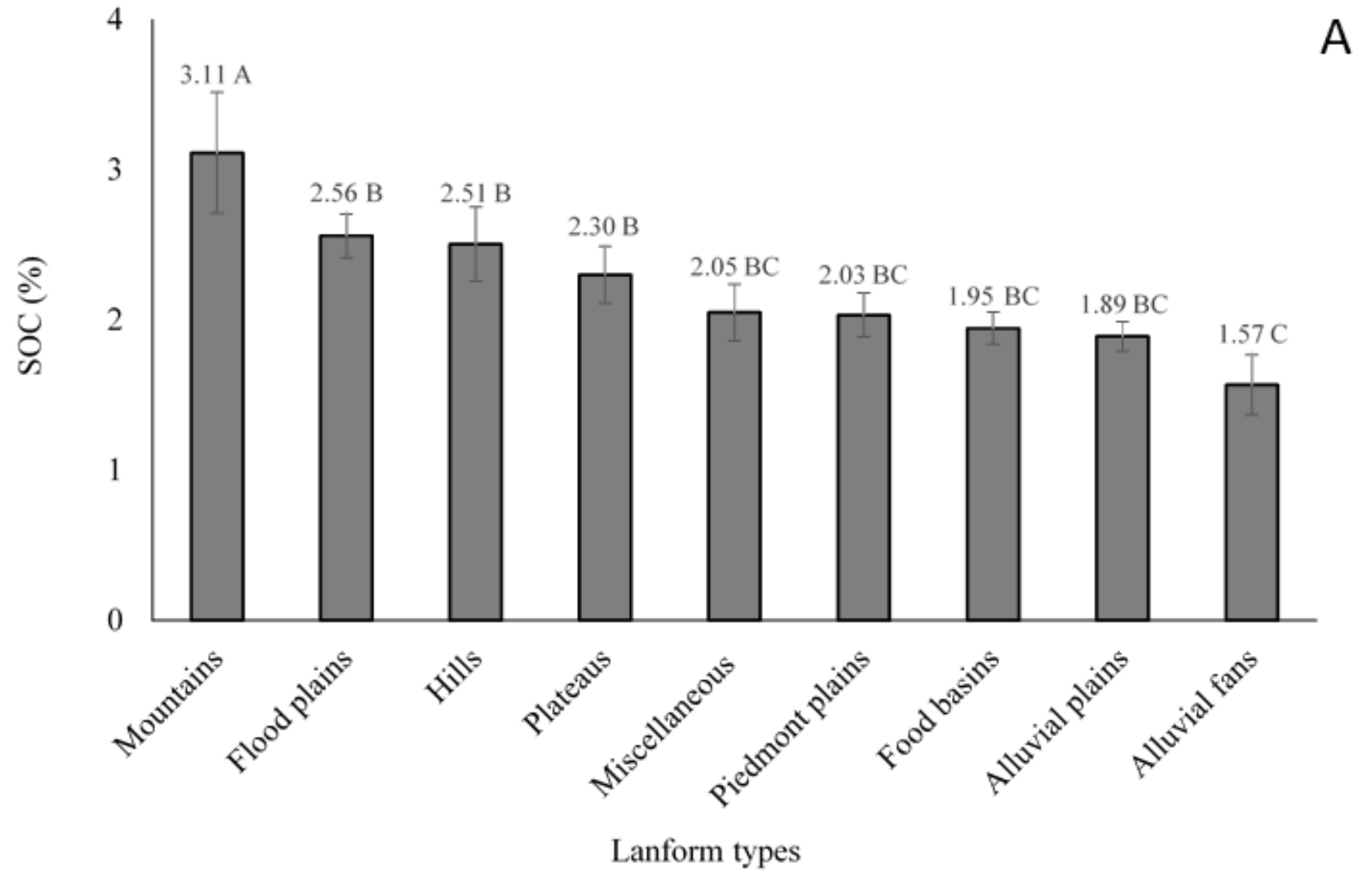

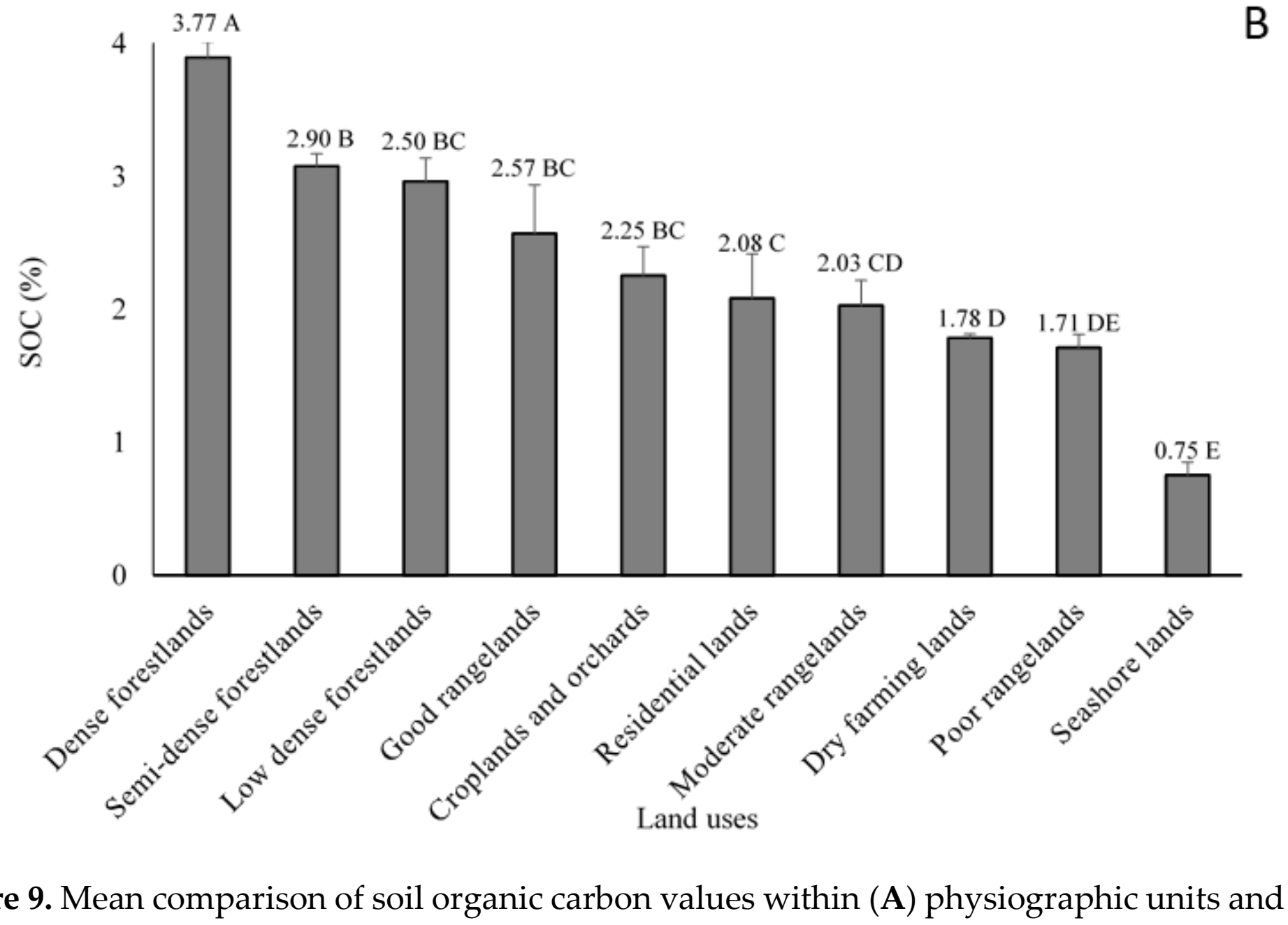

**Figure 9.** Mean comparison of soil organic carbon values within (**A**) physiographic units and (**B**) land uses in the Mazandaran province. Values with different letters in each column indicate significant differences ($p < 0.05$). I indicates the error bar (SD) in all columns.

The disturbed soils in croplands and orchards had significantly lower SOC compared with forests and rangelands except for poor rangelands. Soils in residential, dry farming, poor rangelands, and seashore areas had SOC mean values of 2.08%, 1.78%, 1.71%, and 0.75%, respectively. Unsurprisingly, the dense forestlands have significantly the highest SOC content with mean values of 3.77% (Figure 9) followed by the semidense forestlands (2.90%), low dense forestlands (2.50%), good rangelands (2.57%), and moderate rangelands (2.03%), respectively. Emadi et al. [58] reported that the cultivation of virgin forest and pasturelands in Mazandaran province led to about 35 and 30% reduction of SOC content, respectively. The conversion of forest and rangelands into the croplands induces SOC oxidation whereby the topsoil SOC decreases.

## 4. Conclusions

The objective of this study was to determine a reliable algorithm for predicting the SOC contents in Mazandaran province through consideration of six different ML algorithms and using 105 environmental auxiliary variables derived from terrain attributes, remote sensing, and climatic data. Thirty-five auxiliary predictors were selected by the GA method. Precipitation, NDVI, MODIS day temperature, MrVBF, and land use were the most important predictors. The results show that the DNN algorithm outperformed other ML algorithms in terms of the power of the prediction uncertainty at the province scale demonstrating that DNN is suitable for use as a robust estimator for SOC mapping in Mazandaran province. The SOC was lower in soils under late geological age (Cenozoic era), while it is accumulated in more developed Ultisols and Mollisols with virgin forest and rangelands in udic SMR classes spread across the middle strip of Mazandaran province. The mesic STR class has significantly higher SOC with high vegetation cover and biomass and probably with a lower C decomposition rate. The predicted SOC map could be used as a base-line for further studies and projects related to the C sequestration development both locally in soils of the Mazandaran province and globally at the worldwide scale. Although the DNN algorithm was found to be the best algorithm to map SOC contents more accurately than other studied ML algorithms, the search for optimized spatial interpolation algorithms is still in its early stages in this province. Moreover, further investigation should be conducted to test the potential of other combination algorithms in this province and test the reliability of DNN reliability for other regions in Iran with different climate and agro-ecological structures.

**Author Contributions:** Conceptualization, M.E., and R.T.-M.; methodology, R.T.-M., and T.S.; software, R.T.-M.; validation, M.E., A.M., and T.S.; formal analysis, R.T.-M., and T.S.; investigation, M.E., and R.T.-M.; resources, A.C.; data curation, A.C., and M.D.; writing—original draft preparation, M.E., and R.T.-M.; writing—review and editing, A.M., and T.S.; visualization, R.T.-M.; supervision, M.E., and R.T.-M. All authors have read and agreed to the published version of the manuscript.

**Funding:** This study is financially supported by the Sari Agricultural Sciences and Natural Resources University (SANRU) (under grant no.T161-1368). The work of Ruhollah Taghizadeh-Mehrjardi has been supported by the Alexander von Humboldt Foundation under the grant number: Ref 3.4-1164573-IRN-GFHERMES-P. Thomas Scholten thanks the German Research Foundation (DFG) for supporting this research through the Collaborative Research Center (SFB 1070) 'Resource Cultures' (subprojects Z, S and B02) and the DFG Cluster of Excellence 'Machine Learning—New Perspectives for Science', EXC 2064/1, project number 390727645. We thank Ruth Kerry, Geography Department, Brigham Young University, USA, who thoroughly revised the technical English of the paper.

**Acknowledgments:** This study is financially supported by the Sari Agricultural Sciences and Natural Resources University (SANRU) (under grant no.T161-1368). The work of Ruhollah Taghizadeh-Mehrjardi has been supported by the Alexander von Humboldt Foundation under the grant number: Ref 3.4-1164573-IRN-GFHERMES-P. Thomas Scholten thanks the German Research Foundation (DFG) for supporting this research through the Collaborative Research Center (SFB 1070) 'Resource Cultures' (subprojects Z, S and B02) and the DFG Cluster of Excellence 'Machine Learning—New Perspectives for Science', EXC 2064/1, project number 390727645. We thank Ruth Kerry, Geography Department, Brigham Young University, USA, who thoroughly revised the technical English of the paper.

**Conflicts of Interest:** The authors declare no conflicts of interest.

## Appendix A

**Table A1.** Environmental auxiliary data initially considered for predicting the distribution of SOC.

| No. | Covariates | Definition |
|---|---|---|
| 1 | Aspect | The compass direction of the maximum rate of change |
| 2 | Catchment Slope | Average gradient above flow path |
| 3 | Channel networks base level | The interpolated channel network base level elevations |
| 4 | Convergence Index | It calculates an index of convergence/divergence regarding to overland flow |
| 5 | Cross-Sectional Curvature | The surface normal and a tangent to the contour—perpendicular to maximum gradient direction |
| 6 | Diffuse Insolation | Calculate the diffuse incoming solar radiation |
| 7 | Direct Insolation | Calculate the direct incoming solar radiation |
| 8 | Downslope Curvature | Calculates the local curvature of a cell as sum of the gradients to its neighbor cells |
| 9 | Elevation | Height above sea level (m) |
| 10 | Flow Accumulation | Calculates accumulated flow |
| 11 | Flow Path Length | The distance from any point in the watershed to the watershed outlet |
| 12 | Local Curvature | The degree to which a curve deviates from a straight line |
| 13 | Mass Balance Index | Balance between soil mass deposited and eroded |
| 14 | Multiresolution Ridge-top Flatness Index | Measure of flatness and lowness |
| 15 | Multiresolution Valley Bottom Flatness Index | Measure of flatness and lowness |
| 16 | Normalized Height | Normalized height is defined by slope height and valley depth |
| 17 | Openness (NegOpen) | How wide a landscape can be viewed from any position |
| 18 | Openness (PosOpen) | How wide a landscape can be viewed from any position |
| 19 | Plan curvature | The curvature of a contour line formed by intersecting a horizontal plane with the surface |
| 20 | Relative Slope Position | The position of one point relative to the ridge and valley of a slope |
| 21 | Slope Gradient | Average gradient above flow path |
| 22 | Slope Length | Calculate the length of slope |
| 23 | Slope Length factor | Slope Length and Steepness factor |
| 24 | Topographic Wetness index | Ln (FA/SG) |
| 25 | Total Insolation | Calculate the total incoming solar radiation |
| 26 | Upslope Curvature | The distance weighted average local curvature in a cell's upslope contributing area |
| 27 | Valley Depth | The vertical distance to a channel network base level |
| 28 | Vector Terrain Ruggedness | Measures terrain ruggedness |
| 29 | Vertical distance to channel networks | The altitude above the channel network |
| 30 | Wind Effect | The Wind Effect is a dimensionless index |
| 31 | Blue | Wavelength of 0.450–0.515 μm of Landsat 8 spectral band |
| 32 | Green | Wavelength of 0.525–0.600 μm of Landsat 8 spectral band |
| 33 | Red | Wavelength of 0.630–0.680 μm of Landsat 8 spectral band |
| 34 | Near infrared | Wavelength of 0.845–0.885 μm of Landsat 8 spectral band |
| 35 | Shortwave infrared-1 | Wavelength of 1.560–1.660 μm of Landsat 8 spectral band |
| 36 | Shortwave infrared-2 | Wavelength of 2.100–2.300 μm of Landsat 8 spectral band |
| 37 | Principal Component 1 | The first principal component of Landsat 8 spectral band |
| 38 | Principal Component 2 | The second principal component of Landsat 8 spectral band |
| 39 | Principal Component 3 | The third principal component of Landsat 8 spectral band |
| 40 | TASSELED CAP 1 | The overall brightness of the image |
| 41 | TASSELED CAP 2 | The overall greenness of the image |
| 42 | TASSELED CAP 3 | The overall wetness of the image |
| 43 | Wetness brightness difference index | TASSELED CAP 3/TASSELED CAP 1 |
| 44 | Atmospherically Resistant Vegetation Index | (−0.18 + 1.17 (NIR − RED/NIR + RED)) |
| 45 | Blue-Wide Dynamic Range Vegetation Index | (0.1 × NIR − BLUE)/(0.1 × NIR + BLUE) |
| 46 | Brightness Index | ((RED)2 + (NIR)2)0.5 |
| 47 | Canopy Index | (SWIR-1-GREEN) |
| 48 | Carbonate Index | (RED/GREEN) |
| 49 | Chlorophyll vegetation index | (NIR × RED/(GREEN)0.5 |
| 50 | Clay Index | (SWIR-1/SWIR-2) |
| 51 | Coloration Index | (RED − GREEN/RED + GREEN) |
| 52 | Differenced Vegetation Index | (NIR − RED) |
| 53 | Enhanced Vegetation Index | (NIR − RED)/(NIR + C1 × RED − C2 × BLUE + L) |
| 54 | Ferrous Minerals | (SWIR-1/NIR) |
| 55 | Green Atmospherically Resistant Vegetation Index | (NIR − (GREEN − (BLUE − RED))/(NIR − (GREEN + (BLUE − RED)) |
| 56 | Green Leaf Index | (2 × GREEN − RED − BLUE)/(2 × GREEN + RED + BLUE) |
| 57 | Green Normalized Difference Vegetation Index | (NIR − GREEN/NIR+ GREEN) |
| 58 | Green Vegetation Index | (0.29 × GREEN − 0.56 × RED + 0.6 × SWIR-1 + 0.49 × GREEN) |
| 59 | Green-Blue NDVI | (NIR − (GREEN + BLUE)/NIR + (GREEN + BLUE)) |
| 60 | Green-Red Vegetation Index | (GREEN − RED) |
| 61 | Gypsum index | (SWIR-1 − NIR)/(SWIR-1 + NIR) |
| 62 | Hue Index | (2 × (RED − GREEN − BLUE))/(GREEN − BLUE) |
| 63 | Infrared Percentage Vegetation Index | (NIR/(NIR+RED)) |
| 64 | Iron Oxide | (RED/BLUE) |
| 65 | Leaf Water Content | (SWIR-1/SWIR-2) |
| 66 | Modified Soil Adjusted Vegetation Index | (0.5 × ((2 × (NIR + 1)) − (((2 × NIR) + 1)2 − 8 × (NIR − RED))0.5)) |
| 67 | Near Infrared Ratio | (NIR/RED) |
| 68 | Norm GREEN | (GREEN/(NIR + RED + GREEN)) |
| 69 | Norm NIR | (NIR/(NIR + RED + GREEN)) |
| 70 | Norm RED | (RED/(NIR + RED + GREEN)) |
| 71 | Normalized Based | ((NIR − (BLUE + GREEN)/(NIR + (BLUE + GREEN))) |
| 72 | Normalized Canopy Index | (SWIR-1 − GREEN/SWIR-1 + GREEN) |
| 73 | Normalized Difference Moisture Index | (NIR − SWIR-1)/(NIR + SWIR-1) |
| 74 | Normalized Difference Salinity Index | (RED − NIR)/(RED + NIR) |
| 75 | Normalized Difference Vegetation Index | (NIR − RED)/(NIR + RED) |
| 76 | Perpendicular Vegetation Index | (NIR − r) cos μ − RED × sin μ |
| 77 | Ratio Vegetation Index | (NIR/RED)/(GREEN + RED) |
| 78 | Redness Index | (RED^2/BLUE × GREEN) |
| 79 | Reflectance Absorption Index | (NIR/(RED + SWIR-1)) |
| 80 | Renormalized difference Vegetation Index | (NIR − RED)/((NIR + RED) ^ 1/2) |
| 81 | MODIS Red | Wavelength of 0.620–0.670 μm of MODIS spectral band |
| 82 | MODIS Near Infrared | Wavelength of 0.841–0.876 μm of MODIS spectral band |
| 83 | MODIS Night Temperature | Land Surface Temperature/Emissivity Daily L3 Global 1 km |
| 84 | MODIS Day Temperature | Land Surface Temperature/Emissivity Daily L3 Global 1 km |
| 85 | MODIS Normalized Difference Vegetation Index | (MODIS NIR − MODIS RED)/(MODIS NIR + MODIS RED) |
| 86 | MODIS Brightness Index | ((MODIS RED)2 + (MODIS NIR)2)0.5 |
| 87 | Soil Adjusted Vegetation Index | (1+ L) × (NIR − RED)/(NIR + RED + L) |
| 88 | Specific Leaf Area Vegetation Index | (NIR/RED + SWIR-1) |
| 89 | Stress Related | ((BLUE× GREEN)/RED) |
| 90 | Vegetation Index | (SWIR-2 − SWIR-1/SWIR-2 + SWIR-1) |
| 91 | Annual Precipitation | It is derived from the monthly rainfall values |
| 92 | Precipitation Seasonality (Coefficient of Variation) | It is derived from the monthly rainfall values |
| 93 | Precipitation of Wettest Month | It is derived from the monthly rainfall values |
| 94 | Precipitation of Driest Month | It is derived from the monthly rainfall values |
| 95 | Mean Annual Temperature | It is derived from the monthly temperature values |
| 96 | Mean Annual Wind Speed | It is derived from the monthly wind speed values |
| 97 | Mean Annual Water Vapor Pressure | It is derived from the monthly water vapor pressure values |
| 98 | Mean Annual Actual Evapo-Transpiration | It is derived from the monthly actual evapo-transpiration values |
| 99 | Mean Annual Potential Evapo-Transpiration | It is derived from the monthly potential evapo-transpiration values |
| 100 | Global Aridity Index | It shows the rainfall deficit for potential vegetative growth |
| 101 | Soil Map | Soil and Water Research Institute of Iran |
| 102 | Geology Map | Soil and Water Research Institute of Iran |
| 103 | Land Use Map | Soil and Water Research Institute of Iran |
| 104 | Physiography Map | Soil and Water Research Institute of Iran |
| 105 | Erosion Classes Map | Soil and Water Research Institute of Iran |

## References

1. Edenhofer, O.; Pichs-Madruga, R.; Sokona, Y.; Seyboth, K.; Kadner, S.; Zwickel, T.; Eickemeier, P.; Hansen, G.; Schlömer, S.; von Stechow, C. *Renewable Energy Sources and Climate Change Mitigation: Special Report of the Intergovernmental Panel on Climate Change*; Cambridge University Press: Cambridge, UK, 2011.
2. Adhikari, K.; Hartemink, A.E. Digital Mapping of Topsoil Carbon Content and Changes in the Driftless Area of Wisconsin, USA. *Soil Sci. Soc. Am. J.* **2015**, *79*, 155–164.
3. Lal, R. Soil carbon sequestration to mitigate climate change. *Geoderma* **2004**, *123*, 1–22.
4. Minasny, B.; McBratney, A.B.; Malone, B.P.; Wheeler, I. Digital mapping of soil carbon. In *Advances in Agronomy*; Elsevier: Amsterdam, The Netherlands, 2013; Volume, 118, pp. 1–47.
5. Yang, R.-M.; Zhang, G.-L.; Liu, F.; Lu, Y.-Y.; Yang, F.; Yang, F.; Yang, M.; Zhao, Y.-G.; Li, D.-C. Comparison of boosted regression tree and random forest models for mapping topsoil organic carbon concentration in an alpine ecosystem. *Ecol. Indic.* **2016**, *60*, 870–878.
6. Emadi, M.; Baghernejad, M.; Bahmanyar, M.A.; Morovvat, A. Changes in soil inorganic phosphorous pools along a precipitation gradient in northern Iran. *Int. J. For. Soil Eros.* **2012**, *2*, 143–147.
7. Ogle, S.M.; Paustian, K. Soil organic carbon as an indicator of environmental quality at the national scale: Inventory monitoring methods and policy relevance. *Can. J. Soil Sci.* **2005**, *85*, 531–540.
8. Jenny, H. *Factors of Soil Formation: A System of Quantitative Pedology*; Courier Corporation: North Chelmsford, MA, USA, 1994.
9. Somarathna, P.; Minasny, B.; Malone, B.P. More data or a better model? Figuring out what matters most for the spatial prediction of soil carbon. *Soil Sci. Soc. Am. J.* **2017**, *81*, 1413–1426.
10. Liakos, K.G.; Busato, P.; Moshou, D.; Pearson, S.; Bochtis, D. Machine learning in agriculture: A review. *Sensors* **2018**, *18*, 2674.
11. Padarian, J.; Minasny, B.; McBratney, A.B. Using deep learning for digital soil mapping: A review aided by machine learning tools. *Soil* **2019**, *5*, 79–89.
12. Mahmoudzadeh, H.; Matinfar, H.R.; Taghizadeh-Mehrjardi, R.; Kerry, R. Spatial prediction of soil organic carbon using machine learning techniques in western Iran. *Geoderma Reg.* **2020**, *21*, e00260.
13. McBratney, A.B.; Stockmann, U.; Angers, D.A.; Minasny, B.; Field, D.J. Challenges for soil organic carbon research. In *Soil Carbon*; Springer: Berlin/Heidelberg, Germany, 2014; pp. 3–16.
14. Lamichhane, S.; Kumar, L.; Wilson, B. Digital soil mapping algorithms and covariates for soil organic carbon mapping and their implications: A review. *Geoderma* **2019**, *352*, 395–413.
15. Zhang, G.; Feng, L.; Song, X. Recent progress and future prospect of digital soil mapping: A review. *J. Integr. Agric.* **2017**, *16*, 2871–2885.

16. Wang, B.; Waters, C.; Orgill, S.; Cowie, A.; Clark, A.; Li Liu, D.; Simpson, M.; McGowen, I.; Sides, T. Estimating soil organic carbon stocks using different modelling techniques in the semi-arid rangelands of eastern Australia. *Ecol. Indic.* **2018**, *88*, 425–438.
17. Xiao, J.; Chevallier, F.; Gomez, C.; Guanter, L.; Hicke, J.A.; Huete, A.R.; Ichii, K.; Ni, W.; Pang, Y.; Rahman, A.F. Remote sensing of the terrestrial carbon cycle: A review of advances over 50 years. *Remote Sens. Environ.* **2019**, *233*, 111383.
18. Mishra, U.; Lal, R.; Liu, D.; Van Meirvenne, M. Predicting the spatial variation of the soil organic carbon pool at a regional scale. *Soil Sci. Soc. Am. J.* **2010**, *74*, 906–914.
19. Veronesi, F.; Schillaci, C. Comparison between geostatistical and machine learning models as predictors of topsoil organic carbon with a focus on local uncertainty estimation. *Ecol. Indic.* **2019**, *101*, 1032–1044.
20. Zhang, H.T.; Gao, M.X. The Application of Support Vector Machine (SVM) Regression Method in Tunnel Fires. *Procedia Eng.* **2018**, *211*, 1004–1011.
21. Castaldi, F.; Chabrillat, S.; Chartin, C.; Genot, V.; Jones, A.; van Wesemael, B. Estimation of soil organic carbon in arable soil in Belgium and Luxembourg with the LUCAS topsoil database. *Eur. J. Soil Sci.* **2018**, *69*, 592–603.
22. Malone, B.P.; McBratney, A.; Minasny, B.; Laslett, G. Mapping continuous depth functions of soil carbon storage and available water capacity. *Geoderma* **2009**, *154*, 138–152.
23. Were, K.; Bui, D.T.; Dick, Ø.B.; Singh, B.R. A comparative assessment of support vector regression, artificial neural networks, and random forests for predicting and mapping soil organic carbon stocks across an Afromontane landscape. *Ecol. Indic.* **2015**, *52*, 394–403.
24. Zhao, Z.; Yang, Q.; Benoy, G.; Chow, T.L.; Xing, Z.; Rees, H.W.; Meng, F.-R. Using artificial neural network models to produce soil organic carbon content distribution maps across landscapes. *Can. J. Soil Sci.* **2010**, *90*, 75–87.
25. Taghizadeh-Mehrjardi, R.; Nabiollahi, K.; Kerry, R. Digital mapping of soil organic carbon at multiple depths using different data mining techniques in Baneh region, Iran. *Geoderma* **2016**, *266*, 98–110.
26. Ballabio, C. Spatial prediction of soil properties in temperate mountain regions using support vector regression. *Geoderma* **2009**, *151*, 338–350.
27. Rossel, R.V.; Behrens, T. Using data mining to model and interpret soil diffuse reflectance spectra. *Geoderma* **2010**, *158*, 46–54.
28. Shepherd, K.D.; Walsh, M.G. Development of reflectance spectral libraries for characterization of soil properties. *Soil Sci. Soc. Am. J.* **2002**, *66*, 988–998.
29. Akpa, S.I.; Odeh, I.O.; Bishop, T.F.; Hartemink, A.E.; Amapu, I.Y. Total soil organic carbon and carbon sequestration potential in Nigeria. *Geoderma* **2016**, *271*, 202–215.
30. Gray, J.M.; Bishop, T.F.; Wilson, B.R. Factors controlling soil organic carbon stocks with depth in eastern Australia. *Soil Sci. Soc. Am. J.* **2015**, *79*, 1741–1751.
31. Martin, M.; Wattenbach, M.; Smith, P.; Meersmans, J.; Jolivet, C.; Boulonne, L.; Arrouays, D. Spatial distribution of soil organic carbon stocks in France: Discussion paper. *Biogeosci. Discuss.* **2010**, *7*, 8409–8443.
32. Wang, B.; Waters, C.; Orgill, S.; Gray, J.; Cowie, A.; Clark, A.; Li Liu, D. High resolution mapping of soil organic carbon stocks using remote sensing variables in the semi-arid rangelands of eastern Australia. *Sci. Total Environ.* **2018**, *630*, 367–378.
33. Nabiollahi, K.; Eskandari, S.; Taghizadeh-Mehrjardi, R.; Kerry, R.; Triantafalis, J. Assessing soil organic carbon stocks under land-use change scenarios using random forest models. *Carbon Manag.* **2019**, *10*, 63–77.
34. Zeraatpisheh, M.; Ayoubi, S.; Jafari, A.; Tajik, S.; Finke, P. Digital mapping of soil properties using multiple machine learning in a semi-arid region, central Iran. *Geoderma* **2019**, *338*, 445–452.
35. Webster, R.; Oliver, M.A. *Geostatistics for Environmental Scientists*; John Wiley & Sons: Hoboken, NJ, USA, 2007.
36. Taghizadeh-Mehrjardi, R.; Neupane, R.; Sood, K.; Kumar, S. Artificial bee colony feature selection algorithm combined with machine learning algorithms to predict vertical and lateral distribution of soil organic matter in South Dakota, USA. *Carbon Manag.* **2017**, *8*, 277–291.
37. Hinton, G.E.; Salakhutdinov, R.R. Reducing the dimensionality of data with neural networks. *Science* **2006**, *313*, 504–507.
38. Salakhutdinov, R.; Tenenbaum, J.B.; Torralba, A. Learning with hierarchical-deep models. *IEEE Trans. Pattern Anal. Mach. Intell.* **2013**, *35*, 1958–1971.

39. Kamilaris, A.; Prenafeta-Boldú, F.X. Deep learning in agriculture: A survey. *Comput. Electron. Agric.* **2018**, *147*, 70–90.
40. Song, X.; Zhang, G.; Liu, F.; Li, D.; Zhao, Y.; Yang, J. Modeling spatio-temporal distribution of soil moisture by deep learning-based cellular automata model. *J. Arid Land* **2016**, *8*, 734–748.
41. Padarian, J.; Minasny, B.; McBratney, A. Using deep learning to predict soil properties from regional spectral data. *Geoderma Reg.* **2019**, *16*, e00198.
42. Wadoux, A.M.J.C.; Padarian, J.; Minasny, B. Multi-source data integration for soil mapping using deep learning. *SOIL* **2019**, *5*, 107–119.
43. Xu, Z.; Zhao, X.; Guo, X.; Guo, J. Deep Learning Application for Predicting Soil Organic Matter Content by VIS-NIR Spectroscopy. *Comput. Intell. Neurosci.* **2019**, *2019*, 3563761.
44. Taghizadeh-Mehrjardi, R.; Schmidt, K.; Amirian-Chakan, A.; Rentschler, T.; Zeraatpisheh, M.; Sarmadian, F.; Valavi, R.; Davatgar, N.; Behrens, T.; Scholten, T. Improving the Spatial Prediction of Soil Organic Carbon Content in Two Contrasting Climatic Regions by Stacking Machine Learning Models and Rescanning Covariate Space. *Remote Sens.* **2020**, *12*, 1095.
45. Shirani, H.; Habibi, M.; Besalatpour, A.; Esfandiarpour, I. Determining the features influencing physical quality of calcareous soils in a semiarid region of Iran using a hybrid PSO-DT algorithm. *Geoderma* **2015**, *259*, 1–11.
46. Xie, H.; Zhao, J.; Wang, Q.; Sui, Y.; Wang, J.; Yang, X.; Zhang, X.; Liang, C. Soil type recognition as improved by genetic algorithm-based variable selection using near infrared spectroscopy and partial least squares discriminant analysis. *Sci. Rep.* **2015**, *5*, 10930.
47. Pourmohammadali, B.; Hosseinifard, S.J.; Salehi, M.H.; Shirani, H.; Boroujeni, E. Effects of soil properties, water quality and management practices on pistachio yield in Rafsanjan region, southeast of Iran. *Agric. Water Manag.* **2019**, *213*, 894–902.
48. Besalatpour, A.A.; Ayoubi, S.; Hajabbasi, M.A.; Jazi, A.Y.; Gharipour, A. Feature Selection Using Parallel Genetic Algorithm for the Prediction of Geometric Mean Diameter of Soil Aggregates by Machine Learning Methods. *Arid Land Res. Manag.* **2014**, *28*, 383–394.
49. Behrens, T.; Zhu, A.X.; Schmidt, K.; Scholten, T. Multi-scale digital terrain analysis and feature selection for digital soil mapping. *Geoderma* **2010**, *155*, 175–185.
50. Taghizadeh-mehrjardi, R.; Toomanian, N.; Khavaninzadeh, A.; Jafari, A.; Triantafilis, J. Predicting and mapping of soil particle-size fractions with adaptive neuro-fuzzy inference and ant colony optimization in central I ran. *Eur. J. Soil Sci.* **2016**, *67*, 707–725.
51. Calixto, W.P.; Martins Neto, L.; Wu, M.; Kliemann, H.J.; de Castro, S.S.; Yamanaka, K. Calculation of soil electrical conductivity using a genetic algorithm. *Comput. Electron. Agric.* **2010**, *71*, 1–6.
52. Welikala, R.A.; Fraz, M.M.; Dehmeshki, J.; Hoppe, A.; Tah, V.; Mann, S.; Williamson, T.H.; Barman, S.A. Genetic algorithm based feature selection combined with dual classification for the automated detection of proliferative diabetic retinopathy. *Comput. Med Imaging Graph.* **2015**, *43*, 64–77.
53. Zeraatpisheh, M.; Jafari, A.; Bodaghabadi, M.B.; Ayoubi, S.; Taghizadeh-Mehrjardi, R.; Toomanian, N.; Kerry, R.; Xu, M. Conventional and digital soil mapping in Iran: Past, present, and future. *Catena* **2020**, *188*, 104424.
54. Guan, J.H.; Deng, L.; Zhang, J.G.; He, Q.Y.; Shi, W.Y.; Li, G.; Du, S. Soil organic carbon density and its driving factors in forest ecosystems across a northwestern province in China. *Geoderma* **2019**, *352*, 1–12.
55. Cruz-Cárdenas, G.; López-Mata, L.; Ortiz-Solorio, C.A.; Villaseñor, J.L.; Ortiz, E.; Silva, J.T.; Estrada-Godoy, F. Interpolation of mexican soil properties at a scale of 1:1,000,000. *Geoderma* **2014**, *213*, 29–35.
56. Guo, Z.; Adhikari, K.; Chellasamy, M.; Greve, M.B.; Owens, P.R.; Greve, M.H. Selection of terrain attributes and its scale dependency on soil organic carbon prediction. *Geoderma* **2019**, *340*, 303–312.
57. Emadi, M.; Shahriari, A.R.; Sadegh-Zadeh, F.; Jalili Seh-Bardan, B.; Dindarlou, A. Geostatistics-based spatial distribution of soil moisture and temperature regime classes in Mazandaran province, northern Iran. *Arch. Agron. Soil Sci.* **2016**, *62*, 502–522.
58. Emadi, M.; Baghernejad, M.; Memarian, H.R. Effect of land-use change on soil fertility characteristics within water-stable aggregates of two cultivated soils in northern Iran. *Land Use Policy* **2009**, *26*, 452–457.
59. Zeraatpishe, M.; Khormali, F. Carbon stock and mineral factors controlling soil organic carbon in a climatic gradient, Golestan province. *J. Soil Sci. Plant Nutr.* **2012**, *12*, 637–654.
60. Darabi, N. Mapping Saline Soils Using GIS and RS Techniques. Master of Science M.S. Thesis, Sari University of Agricultural Sciences and Natural Resources, Sari, Iran, 2016.

61. Maldari, M. Testing Performance of Vis-Infrared Spectral Reflectance for Estimation of Soil Properties. Master of Science M.S. Thesis, Sari University of Agricultural Sciences and Natural Resources, Sari, Iran, 2016.
62. Masoudi, S. Using Geostatistical and Fuzzy Approaches for Delineation of Soil Management Zone by Soil Properties and Wheat Yield, Northern Iran. Master of Science M.S. Thesis, Sari University of Agricultural Sciences and Natural Resources, Sari, Iran, 2016.
63. Sajjadi, F. Spatial Variability of Some Soil Properties in Different Landscape, Northern Iran. Master of Science M.S. Thesis, Sari University of Agricultural Sciences and Natural Resources, Sari, Iran, 2016.
64. Sojoodeh, A. Spatial Variability of Some Soil Physical and Chemical Properties and Comparison of Geostatistical Approaches in Soil Mapping. Master of Science M.S. Thesis, Sari University of Agricultural Sciences and Natural Resources, Sari, Iran, 2015.
65. Amiri, E. Calibration and testing of the Aquacrop model for rice under water and nitrogen management. *Commun. Soil Sci. Plant Anal.* **2016**, *47*, 387–403.
66. Sayão, V.M.; Demattê, J.A. Soil texture and organic carbon mapping using surface temperature and reflectance spectra in Southeast Brazil. *Geoderma Reg.* **2018**, *14*, e00174.
67. Gallant, J.C.; Dowling, T.I. A multi-resolution index of valley bottom flatness for mapping depositional areas. *Water Resour. Res.* **2003**, *39*, 1347.
68. Fick, S.E.; Hijmans, R.J. WorldClim 2: New 1-km spatial resolution climate surfaces for global land areas. *Int. J. Climatol.* **2017**, *37*, 4302–4315.
69. Banaei, M.; Moameni, A.; Bybordi, M.; Malakouti, M. *The Soils of Iran: New Achievements in Perception, Management and Use*; Soil and Water Research Institute: Tehran, Iran, 2005.
70. Tajik, S.; Zarinkamar, F.; Soltani, B.M.; Nazari, M. Induction of phenolic and flavonoid compounds in leaves of saffron (*Crocus sativus* L.) by salicylic acid. *Sci. Hortic.* **2019**, *257*, 108751.
71. Huang, Y.; Lan, Y.; Thomson, S.J.; Fang, A.; Hoffmann, W.C.; Lacey, R.E. Development of soft computing and applications in agricultural and biological engineering. *Comput. Electron. Agric.* **2010**, *71*, 107–127.
72. Kuhn, M. Building predictive models in R using the caret package. *J. Stat. Softw.* **2008**, *28*, 1–26.
73. González Costa, J.J.; Reigosa, M.J.; Matías, J.M.; Covelo, E.F. Soil Cd, Cr, Cu, Ni, Pb and Zn sorption and retention models using SVM: Variable selection and competitive model. *Sci. Total Environ.* **2017**, *593–594*, 508–522.
74. Abrougui, K.; Gabsi, K.; Mercatoris, B.; Khemis, C.; Amami, R.; Chehaibi, S. Prediction of organic potato yield using tillage systems and soil properties by artificial neural network (ANN) and multiple linear regressions (MLR). *Soil Tillage Res.* **2019**, *190*, 202–208.
75. Ochoa-Martínez, C.I.; Ayala-Aponte, A.A. prediction of mass transfer kinetics during osmotic dehydration of apples using neural networks. *LWT Food Sci. Technol.* **2007**, *40*, 638–645.
76. Trigui, M.; Gabsi, K.; Amri, I.E.; Helal, A.N.; Barrington, S. Modular Feed Forward Networks to Predict Sugar Diffusivity from Date Pulp Part I. Model Validation. *Int. J. Food Prop.* **2011**, *14*, 356–370.
77. Fernandes, M.M.H.; Coelho, A.P.; Fernandes, C.; da Silva, M.F.; Dela Marta, C.C. Estimation of soil organic matter content by modeling with artificial neural networks. *Geoderma* **2019**, *350*, 46–51.
78. Yilmaz, I.; Kaynar, O. Multiple regression, ANN (RBF, MLP) and ANFIS models for prediction of swell potential of clayey soils. *Expert Syst. Appl.* **2011**, *38*, 5958–5966.
79. Candel, A.; Parmar, V.; LeDell, E.; Arora, A. *Deep learning with H2O*; H2O.ai Inc.: Mountain View, CA, USA, 2016.
80. Breiman, L.; Friedman, J.H.; Olshen, R.A.; Stone, C.J. *Classification and Regression Trees*; Wadsworth International Group: Belmont, CA, USA, 1984; Volume 432, pp. 151–166.
81. Mikkonen, H.G.; van de Graaff, R.; Clarke, B.O.; Dasika, R.; Wallis, C.J.; Reichman, S.M. Geochemical indices and regression tree models for estimation of ambient background concentrations of copper, chromium, nickel and zinc in soil. *Chemosphere* **2018**, *210*, 193–203.
82. Malone, B.P.; Styc, Q.; Minasny, B.; McBratney, A.B. Digital soil mapping of soil carbon at the farm scale: A spatial downscaling approach in consideration of measured and uncertain data. *Geoderma* **2017**, *290*, 91–99.
83. Appelhans, T.; Mwangomo, E.; Hardy, D.R.; Hemp, A.; Nauss, T. Evaluating machine learning approaches for the interpolation of monthly air temperature at Mt. Kilimanjaro, Tanzania. *Spat. Stat.* **2015**, *14*, 91–113.
84. Kuhn, M.; Weston, S.; Keefer, C.; Coulter, N. Cubist models for regression. *R Package Vignette R Package Version 0.0* **2012**, *18*, pp. 223–244

85. Breiman, L. Random Forests. *Mach. Learn.* **2001**, *45*, 5–32.
86. Fu, B.; Wang, Y.; Campbell, A.; Li, Y.; Zhang, B.; Yin, S.; Xing, Z.; Jin, X. Comparison of object-based and pixel-based Random Forest algorithm for wetland vegetation mapping using high spatial resolution GF-1 and SAR data. *Ecol. Indic.* **2017**, *73*, 105–117.
87. Houborg, R.; McCabe, M.F. A hybrid training approach for leaf area index estimation via Cubist and random forests machine-learning. *ISPRS J. Photogramm. Remote Sens.* **2018**, *135*, 173–188.
88. Chen, T.; Guestrin, C. Xgboost: A scalable tree boosting system. In *Proceedings of the 22nd ACM Sigkdd International Conference on Knowledge Discovery and Data Mining*; ACM: New York, NY, USA, 2016; pp. 785–794.
89. Fan, J.; Wang, X.; Wu, L.; Zhou, H.; Zhang, F.; Yu, X.; Lu, X.; Xiang, Y. Comparison of Support Vector Machine and Extreme Gradient Boosting for predicting daily global solar radiation using temperature and precipitation in humid subtropical climates: A case study in China. *Energy Convers. Manag.* **2018**, *164*, 102–111.
90. Li, W.; Fu, H.; Yu, L.; Gong, P.; Feng, D.; Li, C.; Clinton, N. Stacked Autoencoder-based deep learning for remote-sensing image classification: A case study of African land-cover mapping. *Int. J. Remote Sens.* **2016**, *37*, 5632–5646.
91. Sa, I.; Popović, M.; Khanna, R.; Chen, Z.; Lottes, P.; Liebisch, F.; Nieto, J.; Stachniss, C.; Walter, A.; Siegwart, R. Weedmap: A large-scale semantic weed mapping framework using aerial multispectral imaging and deep neural network for precision farming. *Remote Sens.* **2018**, *10*, 1423.
92. Emadi, M.; Baghernejad, M.; Emadi, M.; Maftoun, M. Assessment of some soil properties by spatial variability in saline and sodic soils in Arsanjan plain, Southern Iran. *Pak. J. Biol. Sci.* **2008**, *11*, 238–243.
93. Wang, Y.; Zhang, Z.; Feng, L.; Du, Q.; Runge, T. Combining Multi-Source Data and Machine Learning Approaches to Predict Winter Wheat Yield in the Conterminous United States. *Remote Sens.* **2020**, *12*, 1232.
94. Wang, N.; Zhang, D.; Chang, H.; Li, H. Deep learning of subsurface flow via theory-guided neural network. *J. Hydrol.* **2020**, *584*, 124700.
95. Floody, M.C.; Theng, B.; Reyes, P.; Mora, M. Natural nanoclays: Applications and future trends—A Chilean perspective. *Clay Miner.* **2009**, *44*, 161–176.
96. Mitsa, T. How Do You Know You Have Enough Training Data? 2019. Available online: https://towardsdatascience.com/how-do-you-know-you-have-enough-training-data-ad9b1fd679ee (accessed on 6 June 2020).
97. Zhu, X.; Vondrick, C.; Fowlkes, C.C.; Ramanan, D. Do we need more training data?. *Int. J. Comput. Vis.* **2016**, *119*, 76–92.
98. Nagelkerke, N.J. A note on a general definition of the coefficient of determination. *Biometrika* **1991**, *78*, 691–692.
99. Nickerson, C.A. A note on "A concordance correlation coefficient to evaluate reproducibility". *Biometrics* **1997**, *53*, 1503–1507.
100. Willmott, C.J.; Matsuura, K. Advantages of the mean absolute error (MAE) over the root mean square error (RMSE) in assessing average model performance. *Clim. Res.* **2005**, *30*, 79–82.
101. Minasny, B.; Setiawan, B.I.; Arif, C.; Saptomo, S.K.; Chadirin, Y. Digital mapping for cost-effective and accurate prediction of the depth and carbon stocks in Indonesian peatlands. *Geoderma* **2016**, *272*, 20–31.
102. Griffiths, R.P.; Madritch, M.D.; Swanson, A.K. The effects of topography on forest soil characteristics in the Oregon Cascade Mountains (USA): Implications for the effects of climate change on soil properties. *For. Ecol. Manag.* **2009**, *257*, 1–7.
103. Ma, M.; Chang, R. Temperature drive the altitudinal change in soil carbon and nitrogen of montane forests: Implication for global warming. *Catena* **2019**, *182*, 104126.
104. Falahatkar, S.; Hosseini, S.M.; Ayoubi, S.; Salmanmahiny, A. Predicting soil organic carbon density using auxiliary environmental variables in northern Iran. *Arch. Agron. Soil Sci.* **2016**, *62*, 375–393.
105. Xiong, X.; Grunwald, S.; Myers, D.B.; Kim, J.; Harris, W.G.; Bliznyuk, N. Assessing uncertainty in soil organic carbon modeling across a highly heterogeneous landscape. *Geoderma* **2015**, *251–252*, 105–116.
106. Nabiollahi, K.; Taghizadeh-Mehrjardi, R.; Eskandari, S. Assessing and monitoring the soil quality of forested and agricultural areas using soil-quality indices and digital soil-mapping in a semi-arid environment. *Arch. Agron. Soil Sci.* **2018**, *64*, 696–707.

107. Matsushita, B.; Yang, W.; Chen, J.; Onda, Y.; Qiu, G. Sensitivity of the enhanced vegetation index (EVI) and normalized difference vegetation index (NDVI) to topographic effects: A case study in high-density cypress forest. *Sensors* **2007**, *7*, 2636–2651.
108. Dai, F.; Zhou, Q.; Lv, Z.; Wang, X.; Liu, G. Spatial prediction of soil organic matter content integrating artificial neural network and ordinary kriging in Tibetan Plateau. *Ecol. Indic.* **2014**, *45*, 184–194.
109. Pei, T.; Qin, C.-Z.; Zhu, A.X.; Yang, L.; Luo, M.; Li, B.; Zhou, C. Mapping soil organic matter using the topographic wetness index: A comparative study based on different flow-direction algorithms and kriging methods. *Ecol. Indic.* **2010**, *10*, 610–619.
110. Schillaci, C.; Lombardo, L.; Saia, S.; Fantappiè, M.; Märker, M.; Acutis, M. Modelling the topsoil carbon stock of agricultural lands with the Stochastic Gradient Treeboost in a semi-arid Mediterranean region. *Geoderma* **2017**, *286*, 35–45.
111. Stevens, A.; van Wesemael, B.; Bartholomeus, H.; Rosillon, D.; Tychon, B.; Ben-Dor, E. Laboratory, field and airborne spectroscopy for monitoring organic carbon content in agricultural soils. *Geoderma* **2008**, *144*, 395–404.
112. Gray, J.; Karunaratne, S.; Bishop, T.; Wilson, B.; Veeragathipillai, M. Driving factors of soil organic carbon fractions over New South Wales, Australia. *Geoderma* **2019**, *353*, 213–226.
113. Khormali, F.; Ghergherechi, S.; Kehl, M.; Ayoubi, S. Soil formation in loess-derived soils along a subhumid to humid climate gradient, Northeastern Iran. *Geoderma* **2012**, *179–180*, 113–122.
114. Pourmasoumi, M.; Khormali, F.; Ayoubi, S.; Kehl, M.; Kiani, F. Development and magnetic properties of loess-derived forest soils along a precipitation gradient in northern Iran. *J. Mt. Sci.* **2019**, *16*, 1848–1868.
115. Rossi, A.M.; Rabenhorst, M.C. Organic carbon dynamics in soils of Mid-Atlantic barrier island landscapes. *Geoderma* **2019**, *337*, 1278–1290.